\documentclass[11pt]{article}

\usepackage[preprint]{acl}

\usepackage{times}
\usepackage{latexsym}

\usepackage[T1]{fontenc}

\usepackage[utf8]{inputenc}

\usepackage{microtype}

\usepackage{inconsolata}

\usepackage{graphicx}

\usepackage{multirow}
\usepackage{booktabs}
\usepackage{amsmath}
\usepackage{amsfonts}
\usepackage{stfloats}

\usepackage{xcolor}
\usepackage[breakable]{tcolorbox}
\tcbuselibrary{skins,listings,breakable}

\edef\PBsavedtt{\ttdefault}
\usepackage[scaled=1.0]{sourcecodepro}
\renewcommand{\ttdefault}{\PBsavedtt}
\newcommand{\promptfont}{\fontfamily{SourceCodePro-TLF}\selectfont}
\usepackage{upquote}

\title{Steering Black-Box Large Language Model Agents with \\ Pitfall Anticipation World Model}

\author{Jiongxiao Wang$^{1}$\thanks{Work was done during an internship at AWS AI.} \quad Han Ding$^{2}$ \quad Youzhi Luo$^{2}$ \quad Yawei Wang$^{2}$ \\ \textbf{Yun Zhou}$^{2}$ \quad \textbf{Linbo Liu}$^{2}$ \quad  \textbf{Panpan Xu}$^{2}$ \\
\textsuperscript{1}University of Wisconsin–Madison; \textsuperscript{2}AWS AI \\ 
\small{
\texttt{jwang2929@wisc.edu; \{handing, yzluo, yawenwan, yunzzhou, linbol, xupanpan\}@amazon.com}}
}

\begin{document}
\maketitle
\begin{abstract}
Frontier black-box Large Language Model (LLM) agents are proficient at long-horizon reasoning and precise action generation. However, when it comes to a specific agentic task, they have no internalized knowledge of how that particular environment unfolds. A world model offers a natural remedy, providing the agent with environment specific knowledge by foreseeing how the environment will respond to the action it is about to take. Prior world models for LLM agents mainly predict next observations at each step, which is costly and often uninformative because individual observations provide limited task-relevant guidance and not every step requires call for foresight. This raises a key question: \textbf{what a world model should predict and when should it be invoked for agentic tasks?} Our pilot study suggests that world model guidance is most useful when it anticipates pitfalls within a bounded multi-step future. Motivated by this, we introduce \textbf{Pitfall Anticipation World Model (PAWM)}, which selectively intervenes when a potential pitfall arises and predicts the impending failure to steer the agent away from it. PAWM is trained in two stages: Supervised Fine-Tuning (SFT) on synthetic data labeled with agent pitfalls, followed by GRPO co-training with a black-box LLM to further optimize it in the target environment. PAWM improves Claude Sonnet 4.5 performance by \textbf{24.3\%} on ScienceWorld and \textbf{7.7\%} on WebArena-Lite, outperforming a range of baselines. These results demonstrate that PAWM provides an effective form of world model guidance for black-box LLM agents.
\end{abstract}

\section{Introduction}
Frontier LLMs such as GPT-5 and Claude are increasingly deployed in customized agentic applications~\citep{yao2023react}, from enterprise workflow automation~\citep{yao2024tau, barres2025tau} to specialized domain assistants~\citep{trivedi2024appworld, jimenez2024swe}. Owing to large-scale training, these models already excel at the core competencies such as instruction following, general reasoning, task planning, long contexts processing, and producing precise, well-formatted actions via function calling or code. What they still lack is not general capability but task-specific reliability for agents. Since these models are black-box, such reliability cannot be instilled by fine-tuning their weights. This has inspired growing interest in steering black-box LLM agents~\citep{li2024martryoshka, asawa2025train, lingam2026exectune, liu2026clear}, where an external module injects guidance that shapes the frozen model's behavior at inference time. However, prior steering methods mainly focus on assisting agent planning, such as decomposing the goal into subgoals or laying out high-level action sequences. Advising black-box LLM agents on it reinforces capabilities they already possess, rather than providing the environment knowledge they lack about how a specific agentic environment responds to their actions. We therefore propose to steer black-box LLM agents with world models~\citep{ha2018wm, li2026bridging}, which can supply this missing environment knowledge.


\begin{figure*}[ht]
\centering
\includegraphics[width=0.8\textwidth]{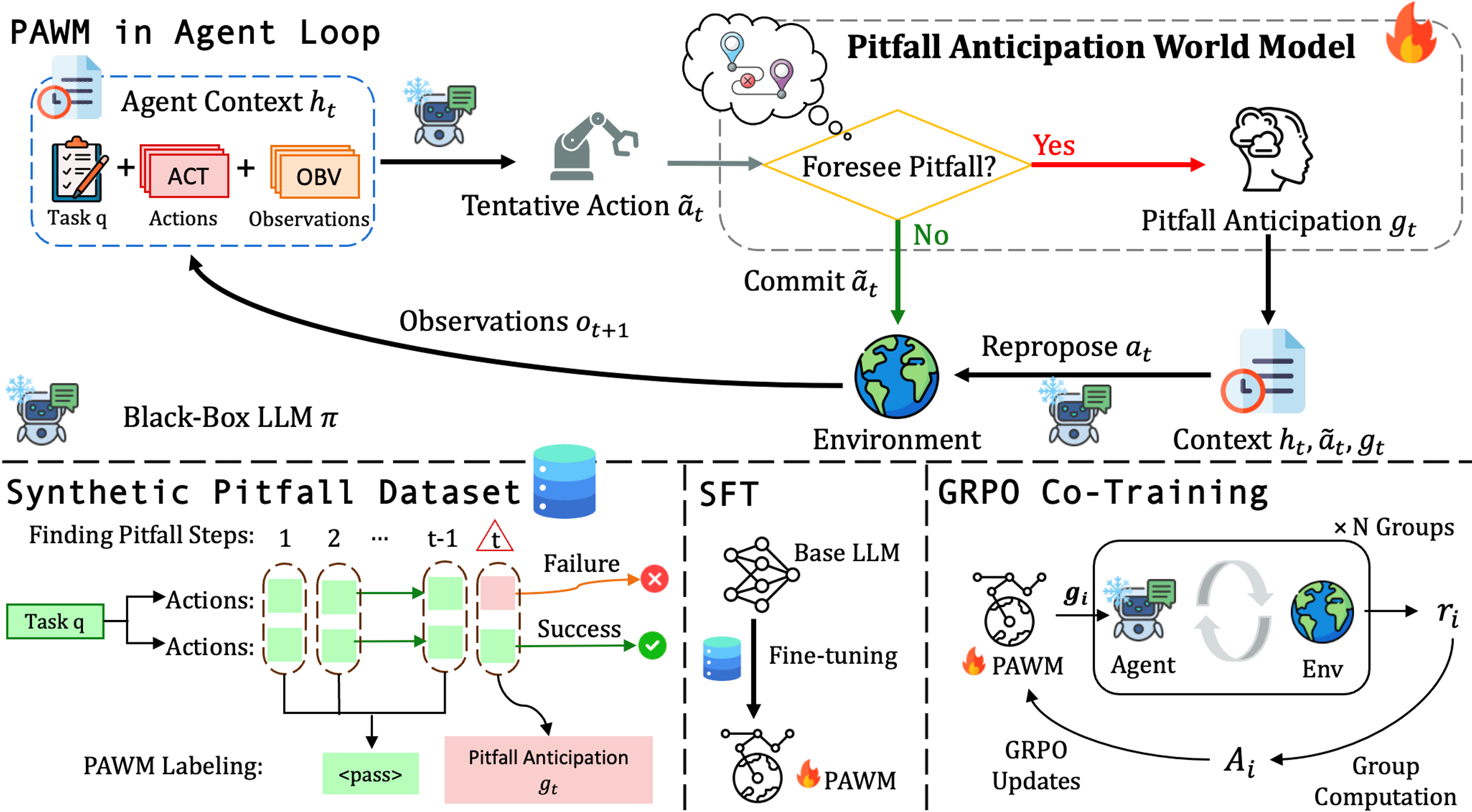}
\caption{Illustration for Pitfall Anticipation World Model in steering black-box LLM agents.}
\end{figure*}

Rather than encoding what to do, a world model captures how the environment behaves, how states transit in response to actions, and how a task trajectory is likely to unfold. A black-box LLM, although capable in general, has never internalized such environment-specific dynamics for the agent task at hand. This makes the world model a natural approach for supplying black-box LLM agents with the grounding they lack. A growing body of work has begun to explore LLM-based world models for agentic tasks, using them to foresee future states and thereby guide action selection. Most of these works, however, employ the world model as a simulator of the environment that rolls out actions to the next observations to support candidates ranking or tree-based search~\citep{zhao2023large, gu2024your, chae2025web, zheng2026can, guan2026computer}. Whether using a world model as a direct source of additional environment knowledge in the agent loop helpful or not remains unclear. For instance, \citet{qian2026current} found that a generative world model provides no consistent benefit for agents, whether invoked adaptively or at every step, with performance sometimes even worse than when acting without a world model.

Therefore, how to effectively deliver a world model's guidance remains an open problem. Two design decisions are central here, and prior work has left both implicit. \textbf{First, what should the world model output?} Environment knowledge can be surfaced at very different granularities, from the immediate consequence of the current step to a summary of the entire remaining trajectory, and it is not obvious which is most useful to a frozen agent. \textbf{Second, when should the world model intervene?} Guiding at every step supplies the most information but risks overwhelming or misdirecting an already-capable agent, whereas intervening too rarely risks staying silent at the steps that matter. Neither question has a settled answer, and together they define the design space for this work.

To answer these questions, we first conduct a systematic pilot study on the ScienceWorld~\citep{wang2022scienceworld} and WebArena-Lite~\citep{zhou2024webarena} benchmarks, in each case training a world model via SFT on collected agent trajectories to steer a Claude Sonnet 4.5 agent model. Along the world model output axis, we compare guidance of increasing horizon: the next observation alone, the segment up to the next subgoal, and the entire remaining trajectory. Our experiment results reveal that guidance bounded at the next subgoal works best, indicating that a world model should look several steps ahead yet stop at a meaningful boundary. Along the intervention axis, we start from the observation that most guidance leaves the agent's action unchanged, and ask what distinguishes the guidance that does change it. Both a lexical and an LLM-based semantic analysis show that pitfall-related generations dominate the guidance that changes the agent's action. To verify that anticipated pitfalls indeed make a good timing for intervention, we run a gating ablation that routes only pitfall-anticipating guidance to the agent, and find that it nearly matches guiding at every step in performance with far less cost. These studies point to two design principles: guidance is most useful when it forecasts \textbf{a multi-step future bounded at a meaningful milestone}, and its value concentrates at \textbf{potential pitfalls}.

Inspired by our pilot study, we propose \textbf{Pitfall Anticipatory World Model (PAWM)}, a world model that steers a black-box LLM agent by anticipating pitfalls only when they are likely to arise. Unlike prior world models that simulate candidate actions or provide guidance at every step, PAWM stays silent while the agent is on a promising trajectory and intervenes only when it foresees an impending failure, at which point it predicts the pitfall ahead and issues a multi-step forward-looking guidance that steers the agent away from the action it is about to take. By stepping in only at these key moments, PAWM avoids unnecessary guidance and acts exactly where the agent would otherwise fail, making it both more efficient and effective.

We obtain PAWM in two stages. First, we perform SFT on a synthetic pitfall dataset in which a separate LLM annotates repeatedly sampled trajectories, contrasting a successful and a failed run of the same task to locate the key pitfall steps and describing the corresponding failure risks. This teaches PAWM both to foresee potential pitfalls and to articulate the guidance needed to avoid them. Furthermore, to optimize the anticipation ability for the agent's downstream task success, we apply GRPO~\citep{shao2024deepseekmath} to co-train PAWM with the frozen black-box model, adapting its steering to the specific agent and environment. Integrating PAWM with a Claude Sonnet 4.5 agent yields success rates of \textbf{78.9\%} on ScienceWorld and \textbf{80.4\%} on WebArena-Lite, significantly outperforming the black-box Sonnet agent, which reaches only 54.6\% and 72.7\% respectively. In addition, compared with a range of baselines, including steering black-box LLM agents with pre-task guidance and world models under various settings, PAWM consistently performs best, demonstrating its effectiveness of steering agents with environment-specific pitfall knowledge.

\section{Related Work}

\noindent \textbf{Steering Black-box LLMs.} Frontier LLMs are often accessible only through APIs, making task-specific parameter updates infeasible. Prior work therefore adapts these models externally, either through prompt optimization~\citep{shin2020autoprompt, fernando2024promptbreeder} or by training auxiliary models that provide input-dependent guidance while keeping the underlying LLM fixed~\citep{li2023guiding}. Recent work extends this paradigm to complex agentic tasks, where lightweight advisors provide high-level guidance to steer the behavior of black-box LLM agents~\citep{li2024martryoshka, asawa2025train, liu2026clear}.

\noindent \textbf{World Models.} World models learn environment dynamics to support planning and decision-making~\citep{ha2018wm, hafner2019learning, hafner2019dream}. For LLM agents, they are commonly used to generate synthetic experience~\citep{gao2025websynthesis, chen2025scaling, ding2026dynaweb}, internalize environment dynamics into agents~\citep{zhang2025agent, yu2025dyna, yu2026reinforcement}, or assist action selection through simulation and search~\citep{zhao2023large, gu2024your, chae2025web, zheng2026can, guan2026computer}. Our setting instead considers a world model that continuously assists a black-box LLM in the agent loop. Although recent work explores similar interaction patterns~\citep{qian2026current, zhang2026self}, they offer no systematic study of how a world model should help the agent. Additional discussion of related works is provided in Appendix~\ref{app:related-work}.

\section{Pilot Study}

Before developing our method, we conduct a pilot study to systematically explore the design space introduced above: \textbf{what a world model should output and when it should intervene}. All experiments use Claude Sonnet 4.5 as a fixed black-box agent, guided by a separately trained world model with the base Qwen3.5-27B, ensuring that any performance differences arise from the world model's guidance rather than changes to the agent itself.

\subsection{Problem Definition and Experiment Setup}
\label{sec:setup}
We formalize the agent task as a \textbf{Partially Observable Markov Decision Process (POMDP)} $\mathcal{M}=(\mathcal{S},\mathcal{A},\mathcal{O},\mathcal{T},\mathcal{R})$, where $\mathcal{S}$ denotes the environment state space, $\mathcal{A}$ the action space, $\mathcal{O}$ the observation space, $\mathcal{T}$ the transition dynamics, and $\mathcal{R}$ the reward function. Given a task $q \sim \mathcal{Q}$, the agent does not directly observe the underlying environment state and instead acts based on its interaction history $h_t=(q,o_1,a_1,\dots,a_{t-1},o_t)$ at step $t$, where $o_t$ and $a_t$ denote the observation received and the action taken at step $t$, respectively. The agent here is a frozen black-box policy $\pi$ that can only be queried but not be updated. Our goal is to learn a world model $\omega_\theta$ that improves the agent's performance without modifying $\pi$.

Concretely, at each step, the agent first proposes a tentative action $\tilde{a}_t \sim \pi(\cdot | h_t)$. Rather than executing it immediately, the world model rolls the action forward and produces guidance $g_t \sim \omega_{\theta}(\cdot | h_t,\tilde{a}_t)$ that anticipates its consequences. The resulting guidance is then appended to the agent's context, allowing the agent to repropose the action as $a_t \sim \pi(\cdot | h_t,\tilde{a}_t,g_t)$, either retaining the tentative action or revising it in light of the predicted outcome. Finally, only $a_t$ is executed, yielding the next observation $o_{t+1}$, after which the process repeats. In this way, the world model steers the agent through $\pi$'s own context rather than selecting actions or searching on its behalf.

Note that both the agent and the world model operate \textbf{turn-wise}: at each step, they are queried with the interaction history $h_t$ reconstructed from scratch rather than accumulated through the agent's message list. We adopt this design for two reasons. First, raw observations can be lengthy, such as a full HTML page in WebArena-Lite. Reconstructing $h_t$ allows us to compress past observations, keeping the context manageable over long horizons. Second, when rebuilding the context, we omit previous world model guidance to prevent its influence from carrying over to subsequent decisions. Prompts used to implement this turn-wise agent are provided in Appendix~\ref{app:prompt}.

We evaluate our approach on two benchmarks: \textbf{ScienceWorld}~\citep{wang2022scienceworld}, a text-based environment that demands long-horizon interactions and \textbf{WebArena-Lite}~\citep{zhou2024webarena}, an online web-navigation environment with rich, dynamic observations. In both, Claude Sonnet 4.5 serves as the black-box LLM agent $\pi$ and we obtain the world model $\omega_\theta$ by collecting agent trajectories, extracting the environment dynamics and then perform SFT on them with the base model Qwen3.5-27B. 

\subsection{What Should the World Model Output?}
\label{sec:what}
We first vary the \emph{content} of the world model's output while keeping its intervention at every step, comparing guidance over increasingly longer horizons. Specifically, we consider the following output types. \textbf{(1) Next observation (Next-Obv):} the world model predicts only the immediate observation resulting from the tentative action $\tilde{a}_t$. \textbf{(2) Future trajectory:} the world model instead summarizes the upcoming trajectory at two different horizons: the segment leading to the next subgoal \textbf{(SubGoal-Traj)}, and the entire remaining trajectory to task completion \textbf{(Full-Traj)}. These output types differ only in how their training targets are constructed. Starting from the same set of collected agent trajectories, we prompt an LLM to generate the corresponding target for each output type, yielding one dataset variant per setting. All variants are then trained under the same SFT setup, with details provided in Appendix~\ref{app:sft}. We report Task Success Accuracy \textbf{(Acc)} averaged over three runs, together with the average Action Change Rate \textbf{(ACR)}, defined as the fraction of guided steps at which the agent revises its tentative action $\tilde{a}_t$, to quantify how strongly the world model's output affects the agent. \textbf{N/A} denotes the baseline without a world model.

\begin{table}[ht]
\centering
\scriptsize
\begin{tabular}{l cc cc}
\toprule
\multirow{2}{*}{\bf WM Output}
  & \multicolumn{2}{c}{\bf ScienceWorld}
  & \multicolumn{2}{c}{\bf WebArena-Lite} \\
\cmidrule(lr){2-3}\cmidrule(lr){4-5}
 & {\bf Acc} & {\bf ACR} & {\bf Acc} & {\bf ACR} \\
\midrule
 N/A & 54.6 & -- & 72.7 & -- \\
 Next-Obv        & 58.5 & \bf20.1 & 73.1 & 15.9 \\
Full-Traj     &  65.6 & 19.4 & 71.1 & \bf24.6 \\
SubGoal-Traj      & \bf66.7 & 18.2 & \bf74.1 & 17.7 \\
\bottomrule
\end{tabular}
\caption{\textbf{Acc(\%)} and \textbf{ACR(\%)} of a frozen agent guided by world models with various outputs.}
\label{tab:pilot-what}
\vspace{-2mm}
\end{table}

Table~\ref{tab:pilot-what} summarizes the results and reveals several clear trends. First, predicting only the next observation is a weak steering signal, yielding marginal gains over the baseline without world models. This may be because a single step rarely reveals enough information to advise the agent. Besides, extending the forecast to the entire remaining trajectory does not solve this and even falls below the baseline on WebArena-Lite, where a long-horizon prediction may accumulate error and disrupt an agent that was already on track. The SubGoal-Traj variant performs best on both benchmarks, indicating that the value of a world model lies in a multi-step forward-looking bounded at a meaningful milestone. Furthermore, Next-Obv on ScienceWorld and Full-Traj on WebArena-Lite attain the highest ACR yet the lowest Acc, showing that frequently revising the agent's tentative action does not ensure improved task accuracy.

\subsection{When Should the World Model Intervene?}
\label{sec:when}
As shown in Table~\ref{tab:pilot-what}, most world model interventions do not alter the agent's tentative actions. This motivates us to avoid intervening at every step, which is both costly and potentially disrupts an already capable agent. To understand \emph{when} world model guidance actually matters, we contrast the world model generations that \emph{do} change the tentative action (\textsc{Changed}) against those leave it unchanged (\textsc{Unchanged}).

\noindent\textbf{Lexical Analysis.} We first perform a lexical analysis of which words distinguish the two groups. Since raw frequency is dominated by generic words, we rank words by the weighted log-odds ratio with uninformative Dirichlet prior~\citep{monroe2008fightin}, which contrasts a word's frequency in the \textsc{Changed} and \textsc{Unchanged} groups, then standardizes it into a $z$-score so that rare terms are down-weighted. To surface environment-invariant terms, we compute this $z$-score separately per benchmark and rank words by the minimum of the two, $z^w_{\min}$, which requires a word to be distinctive in both environments. A higher $z^w_{\min}$ therefore means that the word is over-represented in \textsc{Changed} group on both benchmarks. Formulations to compute $z^w_{\min}$ are given in Appendix~\ref{app:lexical}. As shown in Table~\ref{tab:logodds}, the most distinctive words are associated with anticipated failure or stagnation, including explicit negative outcomes such as \texttt{fails} and \texttt{without}, predictions of an unchanged state such as \texttt{remains}, and contrastive markers such as \texttt{however} that signal a mismatch between the intended and predicted outcomes. This suggests that the agent change its action mainly when the world model anticipates that the tentative action would lead to failures or no progress. Refer to Appendix~\ref{app:lexical} for additional top-k words and word cloud visualizations.

\begin{table}[ht]
\centering
\scriptsize
\begin{tabular}{l r r r}
\toprule
\bf Word & \bf $z_{\min}$ & \bf $z_{\text{Web}}$ & \bf $z_{\text{Sci}}$ \\
\midrule
\texttt{without}   & 34.6 & 34.6 & 38.8 \\
\texttt{remains}   & 21.1 & 42.1 & 42.1 \\
\texttt{description} & 15.8 & 29.5 & 15.8 \\
\texttt{however}     & 15.7 & 15.7 & 19.9 \\
\texttt{fails}     & 14.7 & 14.7 & 48.9 \\
\bottomrule
\end{tabular}
\caption{Top-5 words most distinctive of the \textsc{Changed} group, ranked by $z^w_{\min}$.}
\label{tab:logodds}
\vspace{-2mm}
\end{table}

\noindent\textbf{Semantic Analysis.}
Lexical statistics capture word-level patterns but not their semantic meaning, so we complement this analysis by prompting an LLM (Claude Sonnet 4.5) to make a binary judgment for each world model guidance: whether it anticipates a pitfall, i.e., warns that the tentative action will fail, stall, or not achieve the goal. We apply this classifier to $400$ guidance generations randomly sampled from each group and pooled across both benchmarks. The detailed classifier prompt is provided in Appendix~\ref{app:semantic}. Results show that pitfall anticipations account for $76.8\%$ of the \textsc{Changed} group but only $29.0\%$ of the \textsc{Unchanged} group, a statistically significant difference ($\chi^2=181.1$, $\mathrm{df}=1$, $p<0.001$). This semantic analysis reinforces the lexical evidence, showing that the agent is substantially more likely to revise its tentative action when the world model anticipates pitfalls.

\noindent\textbf{Gating Ablation.} The analysis above shows that pitfall anticipation is associated with behavior change, but it remains unclear whether it can serve as a criterion for deciding when world model guidance should intervene. To investigate this, we use an LLM classifier as a gate on the SubGoal-Traj world model: the world model generates guidance at each step, but the gate determines whether that guidance is passed to the agent based on whether it anticipates a pitfall. We compare three conditions: \textbf{All}, which passes every guidance; \textbf{Pitfall-only}, which passes guidance only when it anticipates a pitfall; and \textbf{Non-pitfall-only}, which only passes the non-pitfall guidance. We additionally report \textbf{Intervention Rate (IR)}, the fraction of steps at which guidance is actually passed to the agent, and \textbf{\#Calls}, the average number of black-box agent queries per task. The gating prompt is provided in Appendix~\ref{app:gating}. As shown in Table~\ref{tab:gating}, \textbf{Pitfall-only} nearly matches \textbf{All} in agent performance while requiring substantially fewer black-box API calls, whereas \textbf{Non-pitfall-only} performs clearly worse despite intervening more often. The remaining gap between \textbf{Pitfall-only} and \textbf{All} is expected: the gate relies on a post-hoc classifier, which may not filter out useful guidance, and the world model is also not explicitly optimized to produce pitfall anticipations that are most helpful for steering the agent. Nevertheless, these results suggest that pitfall anticipation provides a promising criterion for when to intervene, retaining most benefit of the world model guidance at a much lower cost.


\begin{table}[ht]

\centering
\scriptsize
\setlength{\tabcolsep}{2.5pt}
\renewcommand{\arraystretch}{0.92}
\begin{tabular}{lcccccccc}
\toprule
\multirow{2}{*}{\bf Gate}
& \multicolumn{4}{c}{\bf ScienceWorld}
& \multicolumn{4}{c}{\bf WebArena-Lite} \\
\cmidrule(lr){2-5}\cmidrule(lr){6-9}
& {\bf Acc} & {\bf ACR} & {\bf IR} & {\bf Calls}
& {\bf Acc} & {\bf ACR} & {\bf IR} & {\bf Calls} \\
\midrule
All
& 66.7 & 18.2 & 100 & 36.5
& 74.1 & 17.7 & 100 & 20.0 \\
Pitfall-only
& 65.2 & 37.1 & 44.4 & 26.7
& 72.9 & 45.9 & 27.5 & 12.8 \\
Non-pitfall-only
& 58.4 & 4.0 & 52.7 & 28.8
& 70.5 & 7.0 & 75.0 & 16.8 \\
\bottomrule
\end{tabular}
\caption{\textbf{Acc(\%)}, \textbf{ACR(\%)}, \textbf{IR(\%)}, and \textbf{\#Calls} under different guidance gate settings.}
\label{tab:gating}
\vspace{-2mm}
\end{table}

In summary, our pilot study presents two design principles: guidance is most useful when it forecasts \textbf{a multi-step future bounded at a meaningful milestone} and its value concentrates at \textbf{potential pitfalls}.

\section{Pitfall Anticipation World Models}
\label{sec:method}

Inspired by the findings of our pilot study, we propose \textbf{Pitfall Anticipation World Model (PAWM)}. PAWM is a trainable world model that monitors a frozen black-box LLM agent and, whenever it foresees an impending failure, describes the pitfall ahead and issues forward-looking guidance that steers the agent away from the action it was about to take. We first specify our PAWM and then describe the two stages that produce it: SFT on synthetic pitfall dataset and GRPO co-training with the black-box LLM agent.

\subsection{PAWM Definition}
\label{sec:method-setup}
PAWM extends the world model of Section~\ref{sec:setup} with one key mechanism, where the output is \emph{gated}. It emits a special \texttt{<pass>} format output when the proposed action has no anticipated pitfall, in which case no guidance is injected and the agent simply commits to $\tilde{a}_t$. Otherwise, it outputs the guidance $g_t$, which is a description of the impending failure and how to avoid it. Then, the agent reproposes the action as $a_t\sim\pi(\cdot | h_t,\tilde{a}_t,g_t)$. Formally,\[ \small \begin{aligned} g_t & \sim \omega_\theta(\cdot|h_t,\tilde a_t) \in \{\texttt{<pass>}\}\cup\mathcal G,\\[-1pt] a_t &= \begin{cases} \tilde a_t, & g_t=\texttt{<pass>},\\ a_t\sim\pi(\cdot|h_t,\tilde a_t,g_t), & \text{otherwise}. \end{cases} \end{aligned} \] where $\mathcal{G}$ is the space of natural language guidance. The gate operationalizes the ``when'' principle, where intervention happens only at decisive steps. While the content of $g_t$ operationalizes the ``what'', which is a multi-step, forward-looking guidance concentrated at pitfalls.

\subsection{Stage 1: SFT on Synthetic Pitfall Dataset}
\label{sec:method-sft}
The goal of the first stage is to teach PAWM \emph{where} pitfalls may occur and \emph{how} to warn against them.

\noindent\textbf{Trajectory collection.} We first gather a pool of agent trajectories by repeatedly running the black-box LLM agent on the training set three times with 1.0 sampling temperature, retaining both successful and failed ones: failures localize where agents actually go wrong, while successes provide the contrastive ``correct'' continuation. 

\noindent\textbf{Pitfall labeling.} For each trajectory, we use a separate LLM (Claude Sonnet 4.5) as an annotator to identify the key pitfall steps at which the task begins to fail. When both successful and failed trajectories are available for the same task, the annotator contrasts them to locate where the failed run diverges from the successful one. For each identified pitfall step, the LLM annotator then describes the potential pitfalls and provides corrective guidance to avoid it, which we denote as $g_t$. All other steps are labeled \texttt{<pass>}. To improve sample utilization, we also handle tasks without a mixture of successful and failed trajectories. For tasks with only successful trajectories, the annotator identifies residual pitfalls, such as suboptimal but non-fatal detours, so that successful runs do not degenerate into all-pass labels. For tasks with only failed trajectories, we rerun the agent with ground-truth guidance injected to obtain successful trajectories, and then contrast these guided successes against the original failures. In this way, tasks whose sampled trajectories are entirely successful or failed can still provide useful pitfall supervision rather than being discarded.

Because the annotator labels trajectories in hindsight, it may leak future knowledge that PAWM has no way of accessing when deployed. To avoid this, we constrain the LLM annotator to describe each pitfall in general, abstracted terms grounded in what the agent has observed up to that step. This yields a dataset $\mathcal{D}=\{(h_t,\tilde{a}_t,g_t)\}$, pairing each history and tentative action with the target guidance. Refer to Appendix~\ref{app:annotator} for prompt details.

\noindent\textbf{Training objective.} We fine-tune $\omega_\theta$ by minimizing the negative log-likelihood loss on $\mathcal{D}$:
\begin{equation}
\small
\mathcal{L}_{\text{SFT}}(\theta)
= -\,\mathbb{E}_{(h_t,\tilde{a}_t,g_t)\sim\mathcal{D}}
\big[\log \omega_\theta(g_t | h_t,\tilde{a}_t)\big] \nonumber
\end{equation}
Since the majority of steps are labeled \texttt{<pass>}, and training on this natural distribution biases $\omega_\theta$ toward always passing and rarely issuing a guidance. To counter this, we down-sample \texttt{<pass>} steps to an empirically selected \texttt{<pass>}-to-anticipation ratio detailed in Section~\ref{sec:pass-ratio}.

\subsection{Stage 2: GRPO Co-Training with the Black-box LLM Agent}
SFT teaches PAWM \emph{when} to raise a pitfall anticipation and \emph{what} guidance to produce, but it optimizes agreement with the annotator rather than whether the guidance actually helps the agent succeed. We therefore further refine PAWM with GRPO, co-training it in the loop with the black-box LLM agent so that its anticipations are rewarded by their outcome-based reward.

For a task $q$, we sample a group of $K$ trajectories $\{\tau_1,\dots,\tau_K\}$ by running PAWM in the loop with the agent, where $\tau_i=\{g_{i,1},...,g_{i,|\tau_i|}\}$ is the sequence of generated pitfall anticipation guidance in the $i$-th trajectory. We then optimize the PAWM $\omega_\theta$ by maximizing the following objective:\[ \small \begin{aligned} \mathcal{J}_{\mathrm{GRPO}}(\theta) &= \mathbb{E}[q\sim Q,\, \{\tau_i\}_{i=1}^{K}\sim\omega_{\theta_{old}}(g|h,\tilde{a})] \nonumber\\[-2pt] &\quad \frac{1}{K}\sum_{i=1}^{K} \frac{1}{|\tau_i|} \sum_{j=1}^{|\tau_i|} \frac{1}{|g_{i,j}|} \sum_{t=1}^{|g_{i,j}|} \ell_{i,j,t}(\theta), \nonumber\\[-2pt] \ell_{i,j,t}(\theta) &= \min\Big[ \rho_{i,j,t}\hat{A}_i,\, \operatorname{clip}(\rho_{i,j,t},1-\epsilon,1+\epsilon) \hat{A}_i \Big] \nonumber\\[-2pt] &\quad -\beta D_{\mathrm{KL}} \big(\omega_\theta\|\omega_{\mathrm{ref}}\big), \nonumber \end{aligned} \] where $\rho_{i,j,t} = \frac{\omega_\theta(g_{i,j,t}|h_{i,j},\tilde{a}_{i,j},g_{i,j,<t})}{\omega_{\theta_{old}}(g_{i,j,t}|h_{i,j},\tilde{a}_{i,j},g_{i,j,<t})}$ is the importance sampling term, $\epsilon$ the clip ratio and $\beta$ the ratio of KL penalty between policy model $\omega_\theta$ and reference model $\omega_{ref}$. Following GRPO, we compute the group-relative advantage from the outcome rewards $\mathbf{r}=\{r_1,\ldots,r_K\}$ as $\hat{A}_{i,t}=\frac{r_i-\text{mean}(\mathbf{r})}{\text{std}(\mathbf{r})}$.


\section{Experiment}
\label{sec:exp}
In this section, we detail the experiments conducted to evaluate PAWM. Specifically, we outline the experimental setup, discuss the main results with additional ablation study.

\begin{table*}[!hb]

\setlength{\tabcolsep}{16pt}
\renewcommand{\arraystretch}{1.15}
\small
\begin{center}
\begin{tabular}{l c@{\hspace{12pt}}c c@{\hspace{12pt}}c}
\toprule
\multirow{2}{*}{\bf Steering Method}
  & \multicolumn{2}{c}{\bf ScienceWorld}
  & \multicolumn{2}{c}{\bf WebArena-Lite} \\
\cmidrule(lr){2-3}\cmidrule(lr){4-5}
  & {\bf Acc} & {\bf Pass@3} & {\bf Acc} & {\bf Pass@3} \\
\midrule
N/A                       & $54.6_{\pm1.9}$   & $66.8$      & $72.7_{\pm0.6}$      & $80.0$     \\
RAG                          & $70.2_{\pm1.4}$     & $78.6$     & $73.3_{\pm2.1}$      & $81.8$     \\
Pre-Task Guidance            & $69.2_{\pm1.9}$      & $79.3$      & $74.3_{\pm1.3}$      & $81.8$      \\
Next-Obv WM                  & $58.5_{\pm1.7}$   & $71.6$     & $73.1_{\pm0.9}$    & $81.2$     \\
Full-Traj WM                 & $65.6_{\pm1.1}$      & $80.1$     & $71.1_{\pm2.1}$      & $80.6$      \\
SubGoal-Traj WM                 & $67.6_{\pm1.6}$      & $79.0$     & $74.3_{\pm2.1}$      & $81.8$      \\
\midrule
\textbf{PAWM (Ours)}     & $\underline{71.4_{\pm1.7}}$       & $\underline{84.5}$     & $\underline{75.0_{\pm1.8}}$      & $\underline{85.5}$      \\
\quad \textbf{+ GRPO}     & $\bf78.9_{\pm0.9}$ & $\bf87.5$  & $\bf80.4_{\pm1.3}$ & $\bf86.1$ \\
\bottomrule
\end{tabular}
\end{center}
\caption{\textbf{Acc(\%)} and \textbf{Pass 3(\%)} of the same Claude Sonnet~4.5 agent under different steering methods, on WebArena-Lite and ScienceWorld. The best and second-best results are highlighted in \textbf{bold} and \underline{underline}, respectively.}
\label{tab:main}
\end{table*}

\subsection{Experimental Setup}
\label{sec:exp-setup}

\textbf{Datasets.} We conduct experiments on two agentic dataset \textbf{WebArena} and \textbf{ScienceWorld}. \textbf{WebArena} comprises 812 tasks. Following \citet{qi2025webrl}, we use 165 verified tasks for evaluation (also named \textbf{WebArena-Lite}) and the remaining 647 tasks as the training set. Since RL rollouts with WebArena issue write actions that mutate website state and can contaminate other tasks, we use WebServ~\citep{lu2025webserv}, which leverages Incus containers to create an isolated container per task. Detailed RL environment settings of WebArena are shown in Appendix~\ref{app:webserv}. For \textbf{ScienceWorld}, it spans 30 task types, each with variations that ScienceWorld partitions into disjoint train/dev/test splits. We collect training data on the train split with up to 30 variations per task (708 tasks) and evaluate on the test split with up to 10 variations per task (271 tasks). 


\noindent\textbf{Models.} Unless otherwise noted, the black-box LLM for the agent is \textbf{Claude Sonnet4.5}, accessed through the Amazon Bedrock API only, and all world models are built on a \textbf{Qwen3.5-27B}\citep{qwen3.5} backbone. 

\noindent\textbf{Training Details.} To obtain PAWM, we first run SFT on the synthetic pitfall dataset and GRPO co-training is then initialized from this checkpoint. Training details are presented in Appendix~\ref{app:training}.

\noindent\textbf{Evaluation Details.} To reduce evaluation variance, we set the temperature of both the agent model and the world model to $0$ during evaluation. Even so, non-determinism environment and minor variability in the LLM serving stack prevent fully reproducible generations. We therefore run every evaluation three times and report averaged results. For both datasets, we report the average Task Success Accuracy (\textbf{Acc}) with its standard deviation, as well as the \textbf{Pass@3}, where a task is deemed solved if at least one of these three runs succeeds.

\noindent\textbf{Baseline Methods.} We compare the PAWM-steered black-box LLM agent against various baselines. \textbf{(1) N/A:} the black-box LLM agent acts on its own without any external guidance. \textbf{(2) RAG:} an embedding model retrieves the most similar previously solved task, and its successful trajectory is inserted into the agent's context as guidance. \textbf{(3) Pre-Task Guidance:} Qwen3.5-27B is trained to generate task-specific guidance conditioned on the task query, providing advice to the agent before it starts. \textbf{(4) Next-Obv/Full-Traj/SubGoal-Traj WM:} We also compare world model steering methods with different output contents. Implementation details for baselines are included in Appendix~\ref{app:baseline}.

\subsection{Results}
We present our main results in Table~\ref{tab:main}, comparing PAWM against various baseline steering methods applied to the same black-box LLM agent Claude Sonnet 4.5. From the table, we can clearly observe that \textbf{PAWM (+GRPO)} performs best across both benchmarks and metrics, reaching 78.9\% Acc on ScienceWorld and 80.4\% on WebArena-Lite, improving over the unassisted agent (\textbf{N/A}) by 24.3\% and 7.7\%. We also observe that GRPO improves Acc far more than Pass@3. This suggests that the two stages play complementary roles, where SFT provides the environment knowledge that sets the ceiling reflected in Pass@3, and GRPO makes the agent realize it more reliably.

\begin{figure*}[!ht]
\centering
\includegraphics[width=0.75\textwidth]{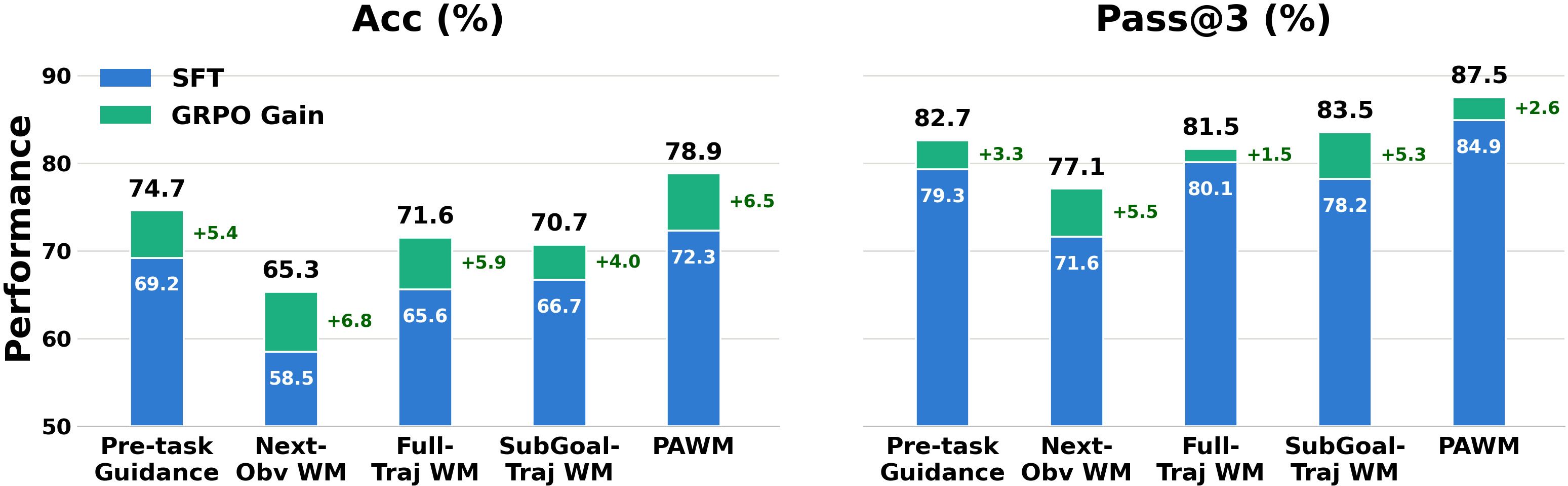}
\caption{GRPO improves over SFT across various steering methods on ScienceWorld.}
\label{fig:grpo}
\vspace{-2mm}
\end{figure*}

For a fair comparison, we also apply GRPO to the baseline steering methods on ScienceWorld, initialized from their SFT checkpoints, including Pre-Task Guidance and world model variants with different outputs. On WebArena-Lite, we run GRPO only on PAWM due to the substantially higher cost and slower online interaction of the environment. As shown in Figure~\ref{fig:grpo}, PAWM consistently achieves the best Acc and Pass@3 among all baselines after GRPO. We further provide qualitative examples in Appendix~\ref{app:example}, illustrating how PAWM steers the agent away from an impending pitfall.

\begin{figure*}[!ht]
\centering
\includegraphics[width=0.75\textwidth]{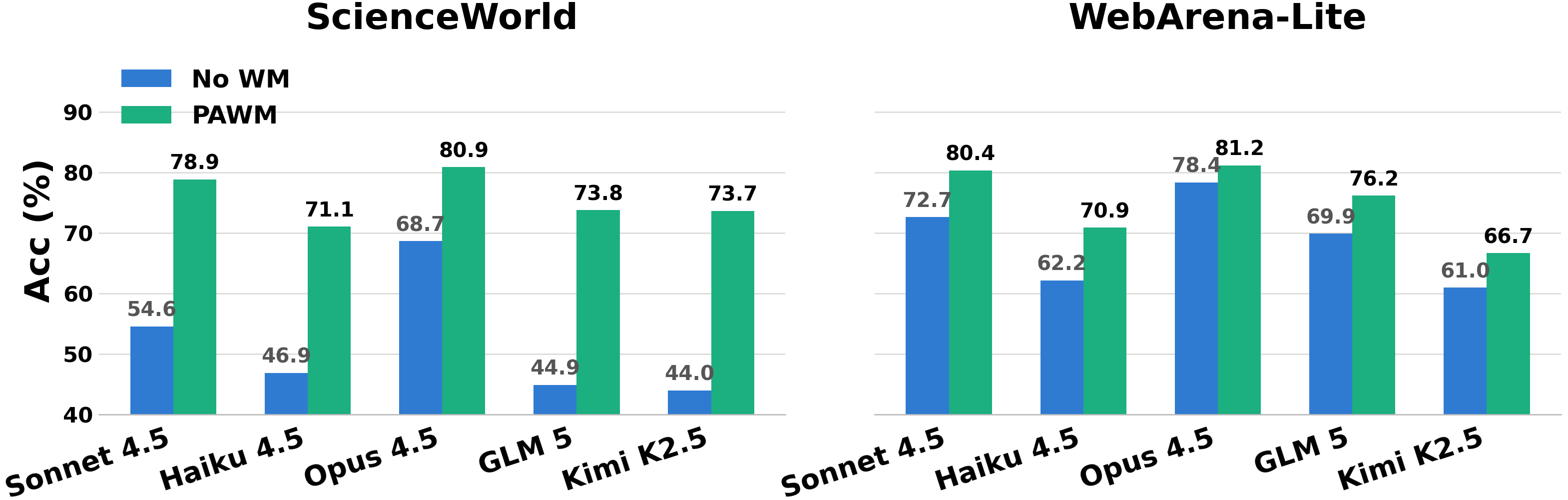}
\caption{Comparison of PAWM with its steered agent under various backbone models.}
\label{fig:transfer}
\vspace{-2mm}
\end{figure*}

\label{sec:pass-ratio}
\noindent\textbf{\texttt{<pass>}-to-anticipation Ratio.} Performing SFT directly on the raw synthetic pitfall dataset, where \texttt{<pass>} steps far outnumber anticipations, risks driving PAWM to stay silent everywhere. We therefore train on datasets balanced to different \texttt{<pass>}-to-anticipation ratios and report Acc, Pass@3, and Intervention Rate (IR) for the resulting SFT models. \textbf{orig} denotes the original ratio without balance. Table~\ref{tab:ratio} shows that a larger \texttt{<pass>} ratio in training consistently leads to a lower IR at evaluation, as the model inherits the label prior of its training distribution. The best accuracy is reached at a ratio of $1.0$ on ScienceWorld and $0.75$ on WebArena-Lite, and we adopt these checkpoints as our final SFT models and as the initialization for GRPO co-training.

\begin{table}[!ht]
\scriptsize
\begin{center}
\begin{tabular}{l ccc ccc}
\toprule
\multirow{2}{*}{\bf Ratios}
  & \multicolumn{3}{c}{\bf ScienceWorld}
  & \multicolumn{3}{c}{\bf WebArena-Lite} \\
\cmidrule(lr){2-4}\cmidrule(lr){5-7}
  & {\bf Acc} & {\bf Pass@3} &  {\bf IR} & {\bf Acc} & {\bf Pass@3} & {\bf IR} \\
\midrule
\bf 0.5 & $69.9$   & $81.6$      & $47.0$      & $72.3$ & $84.2$ & $39.2$    \\
\bf 0.75 & $70.0$     & $82.7$     & $32.8$      & $\bf75.0$ & $\bf85.5$ & $31.0$    \\
\bf 1.0 &  $\bf72.3$      & $\bf84.9$      & $26.4$      & $72.5$ & $83.0$ & $24.1$      \\
\bf 1.5 & $72.1$   & $81.6$     & $19.8$    & $72.5$ & $81.2$ & $26.7$     \\
\bf 2.0 & $69.8$      & $83.8$     & $19.0$      & $71.9$ & $82.4$ & $20.1$      \\
\bf orig & $67.2$      & $78.6$     & $3.2$      & $69.9$ & $81.8$ & $14.6$      \\
\bottomrule
\end{tabular}
\end{center}
\caption{PAWM steered agent performance under different \texttt{<pass>}-to-anticipation Ratios.}
\label{tab:ratio}
\vspace{-2mm}
\end{table}

\noindent\textbf{PAWM Transferability.} It is still unclear whether the environment and task specific knowledge PAWM learns with Claude Sonnet 4.5 transfers to other agent backbones. We therefore conduct additional experiments by directly integrating PAWM (+GRPO) with a range of black-box LLMs, including other size Claude models, Haiku 4.5 and Opus 4.5, as well as open-source models such as GLM 5~\citep{zeng2026glm} and Kimi K2.5~\citep{team2026kimi}. As shown in Figure~\ref{fig:transfer}, PAWM consistently improves accuracy across all tested backbones, demonstrating that the pitfall knowledge it learns is not tied to the agent it was co-trained with.

\section{Conclusion}
This paper studies how to steer black-box LLM agents by providing the environment knowledge they lack with world models. Guided by a pilot study on what a world model should output and when it should intervene, we propose \textbf{Pitfall Anticipation World Model (PAWM)}, which anticipates pitfalls for the agents only when they are likely to arise. PAWM is first trained by SFT on a synthetic pitfall dataset, then co-trained with the frozen agent via GRPO to adapt to the specific environment. Our experiments show that Claude Sonnet 4.5 steered by PAWM outperforms various baselines, and its learned pitfall knowledge transfers across agent backbones, demonstrating that anticipating pitfalls is an effective way for a world model to assist black-box LLM agents.

\section*{Limitations}
Due to the accessibility and cost of repeatedly querying proprietary models, our experiments are primarily scoped to Claude Sonnet 4.5 as the black-box LLM agent used for trajectory collection and GRPO co-training. Although the resulting PAWM transfers to several other agent backbones during evaluation, we do not independently co-train with each of them, so agent specific differences in optimal pitfall anticipation remain underexplored. Our study is also limited to ScienceWorld and WebArena-Lite, and the high cost of online interaction, particularly in WebArena-Lite, limits the scale of reinforcement learning experiments.

In addition, the world models studied here are task-specific: they are trained on trajectories collected from a particular agentic environment rather than as large-scale, general-purpose world models. Accordingly, our conclusions should be interpreted within this task-specific setting, and may not directly extend to broader world models trained across many environments or domains.

Finally, PAWM relies on synthetic pitfall supervision generated by LLM hindsight annotations. Although we constrain the annotations to information available up to the current step, they may still contain noise or teacher-model bias, and our definition of a pitfall may not capture all forms of suboptimal behavior.

\section*{Ethical Considerations}
This work aims to improve the reliability of black-box LLM agents through world model guidance. However, inaccurate guidance may lead to undesirable agent behavior, and the proposed method could potentially be misused to enhance agent capabilities in unintended applications. Our evaluation is limited to benchmark environments, and caution is warranted when deploying the method in real-world settings.

\bibliography{custom}

\clearpage

\appendix
\newtcblisting{promptbox}[2][]{
breakable,
enhanced jigsaw,
colback=gray!3,
colframe=gray!55,
boxrule=0.5pt,
arc=2pt,
left=6pt, right=6pt, top=6pt, bottom=6pt,
title=#2,
fonttitle=\bfseries\small,
coltitle=black,
colbacktitle=gray!18,
listing only,
listing options={
basicstyle=\promptfont\scriptsize\color{black!95},
moredelim={[is][\color{red}]{@@}{@@}},
breaklines=true,
breakatwhitespace=true,
breakautoindent=false,
breakindent=0pt,
columns=fullflexible,
keepspaces=true,
showstringspaces=false,
aboveskip=0pt, belowskip=0pt,
},
#1
}

\section{Related Works}
\label{app:related-work}
\textbf{Steering Black-box LLMs.}
Frontier LLMs such as GPT-5 and Claude are often accessed only through APIs, making their weights unavailable for task-specific adaptation. One line of work adapts them through automatic prompt optimization, searching for a fixed prompt that is reused across all inputs~\citep{shin2020autoprompt, fernando2024promptbreeder}. Such static prompts, however, cannot adapt to individual tasks. A more flexible paradigm is to use a separate model to provide adaptive guidance, an idea long explored in reinforcement learning through teacher-student action advising and policy transfer~\citep{fachantidis2017learning, deng2025energy}. In the context of black-box LLMs, this idea has been instantiated by training a smaller model that produces per-task guidance while leaving the target LLM untouched~\citep{li2023guiding}. Recently, a growing body of works extend this idea to agentic tasks, training a lightweight advisor model to steer a black-box LLM agent with pre-task high-level guidance~\citep{li2024martryoshka, asawa2025train, lingam2026exectune, liu2026clear}. These advisor models, however, mostly supply agent planning or task decomposition, which still builds on the black-box model's general reasoning ability rather than injecting knowledge it does not have. In contrast, we formulate our advisor as a world model providing the environment-specific world knowledge that the agent model lacks.

\noindent \textbf{World Models.}
World models, which predict how an environment evolves in response to actions, were first developed for perception-rich control domains, where latent dynamics models are learned from visual observations and used for planning or policy learning~\citep{ha2018wm, hafner2019learning, hafner2019dream}. With the rise of LLM agents, this paradigm has naturally extended to text-based environments, where LLMs themselves can serve as world models by predicting future states or observations. Existing approaches differ primarily in how these world models are constructed and integrated into agent learning and decision-making. One line of work uses a learned world model as a source of synthetic experience: rather than collecting trajectories from the real environment, which can be costly or impractical, the agent rolls out the world model to generate training data~\citep{gao2025websynthesis, chen2025scaling, ding2026dynaweb}. Another line internalizes environment dynamics directly into the agent model, enabling it to reason about the anticipated consequences of its actions before execution~\citep{zhang2025agent, yu2025dyna, yu2026reinforcement}. More recent work further explores jointly evolving the world model and the agent during training~\citep{liu2026imagine, liu2026comap}.

Our setting instead focuses on using a world model to assist a black-box LLM agent whose parameters cannot be fine-tuned. Prior approaches in this setting typically use the world model to simulate candidate actions and select the most promising one, or to support tree-based search over possible action sequences~\citep{zhao2023large, gu2024your, chae2025web, zheng2026can, guan2026computer}. Such candidate-selection and search-based schemes can be computationally expensive, as the world model must be queried repeatedly across candidates and search branches, while most simulated trajectories are ultimately discarded. We instead consider an alternative paradigm in which the world model remains directly involved in the agent loop and provides guidance during execution, rather than serving only as a simulator for ranking or pruning candidate actions. Although several recent works have begun to explore this form of world-model assistance~\citep{qian2026current, zhang2026self}, they generally adopt a fixed interaction pattern without systematically examining what information the world model should provide or when its guidance is most useful. Our work takes a step toward addressing these questions.

\section{Prompts for Turn-wise Agent}
\label{app:prompt}
This section presents prompts used to realize the turn-wise agent on ScienceWorld and WebArena. Note that the world model only supplies additional guidance, injected at the point where the agent re-decides on its tentative action, leaving the agent's main process untouched.

\subsection{ScienceWorld}
The following shows the system and user prompts for the ScienceWorld agent. Text highlighted in red within braces marks placeholders that are filled with step-specific content at inference time. We also present the format of the agent history, which records the observations and actions from previous steps but excludes the valid action list, as repeating this long list for every past step would consume substantial context. At each step, ScienceWorld environment provides the agent with the \textbf{current location}, the room the agent is in and the objects it contains with their states, the \textbf{current inventory} of objects the agent is carrying, the \textbf{current observation} returned by the environment for the last action, and the \textbf{action list} available at this step for the agent to choose from.

Here we also specify the action reproposal prompt used for world model involved settings. Labels in square brackets, such as \textbf{[Next-Obv]}, \textbf{[Full-Traj]}, and \textbf{[SubGoal-Traj]}, indicate the corresponding experimental setting and are not part of the actual prompt. The text following each label is the prompt content appended after the user prompt to ask the agent to reconsider its proposed action based on the world model guidance.

\begin{promptbox}{System Prompt}
You are an agent playing ScienceWorld, a text-based simulation of elementary-school science experiments. You control a character in a house with rooms (kitchen, bathroom, workshop, greenhouse, etc.) that contain objects you can inspect and manipulate to complete a scientific task.

Rules:
- Each turn you receive the task, an AGENT HISTORY of your prior steps, your current location, your inventory, the result of your last action, and a list of VALID ACTIONS you may take right now.
- You MUST choose exactly one action from the VALID ACTIONS list, copied verbatim. Do not invent objects or actions that are not listed. If the object you want (e.g. a substance for "focus on X") is not in the list, it is not reachable from here yet - move or inspect to bring it into view first, rather than naming it anyway.
- The list of valid actions changes each turn depending on where you are and what is nearby; move around and inspect objects to make more actions available.
- The VALID ACTIONS are shown compactly, grouped by verb to save space:
    * "open: doorA, doorB" means each object is a separate action ("open doorA", "open doorB").
    * "move X to Y  |  X: a, b  |  Y: c, d" is a two-object template: pick one item from X and one from Y to form the action ("move a to c"). Only combinations the game currently allows are valid.
    * A line like "(connect ... - N electrical combos hidden)" means those actions exist but are omitted here because they are not needed for this task.
  Expand a group into a single concrete action and copy that exact wording.

On EVERY step, respond in EXACTLY this two-line format and nothing else:
Think: <your reasoning about the current state, what you have already tried (see AGENT HISTORY), and what to do next>
Action: <a single action copied verbatim from the VALID ACTIONS list>
\end{promptbox}

\begin{promptbox}{User Prompt}
TASK: @@{task instruction}@@

AGENT HISTORY (your prior steps, each with the location (diffed while you stay in a room), inventory, action, and resulting observation):
@@{agent history}@@

CURRENT LOCATION:
@@{current location}@@

CURRENT INVENTORY:
@@{current inventory}@@

CURRENT OBSERVATION:
@@{current observation}@@

VALID ACTIONS (grouped by verb; pick ONE concrete action):
@@{action list}@@

Now select exactly one action from the list above and copy it verbatim as your ACTION.
\end{promptbox}

\begin{promptbox}{Agent History}
--- Step 1 ---
LOCATION: @@{step 1 location}@@
INVENTORY: @@{step 1 inventory}@@
ACTION: @@{step 1 action}@@
OBSERVATION: @@{step 1 observation}@@

--- Step 2 ---
LOCATION: @@{step 2 location}@@
INVENTORY: @@{step 2 inventory}@@
ACTION: @@{step 2 action}@@
OBSERVATION: @@{step 2 observation}@@

...
\end{promptbox}

\begin{promptbox}{Action Reproposal Prompt after User Prompt}
[Next-Obv]
PROPOSED ACTION (the action you are currently about to take):
@@{proposed next action}@@

PREDICTED NEXT OBSERVATION (a world model rolled your PROPOSED ACTION forward one step and predicts the state you would then be looking at - a forecast of the resulting observation, not an instruction):
@@{world model output}@@

Use this PREDICTED NEXT OBSERVATION to sanity-check your PROPOSED ACTION: judge for yourself, using the current state, whether landing on that predicted state is the progress you intend. If the prediction shows your action would fail, have no effect, or lead somewhere unproductive, pick a better action; if the predicted state looks like real progress, keep your PROPOSED ACTION and proceed. Respond in the required format with exactly one action.

[Full-Traj]
PROPOSED ACTION (the action you are currently about to take):
@@{proposed next action}@@

ANTICIPATION (a forecast from a world model that rolled your PROPOSED ACTION forward to the end of the task):
@@{world model output}@@

Use this ANTICIPATION to sanity-check your PROPOSED ACTION: judge for yourself, using the current state, whether landing on that predicted state is the progress you intend. If the anticipation shows your action would fail, stall, or lead somewhere unproductive, pick a better action; if the predicted path looks like real progress, keep your PROPOSED ACTION and proceed. Respond in the required format with exactly one action.

[SubGoal-Traj]
PROPOSED ACTION (the action you are currently about to take):
@@{proposed next action}@@

ANTICIPATION (a forecast from a world model that rolled your PROPOSED ACTION forward to the next milestone):
@@{world model output}@@

Use this ANTICIPATION to sanity-check your PROPOSED ACTION: judge for yourself, using the current state, whether landing on that predicted state is the progress you intend. If the anticipation shows your action would fail, stall, or lead somewhere unproductive, pick a better action; if the predicted path looks like real progress, keep your PROPOSED ACTION and proceed. Respond in the required format with exactly one action.

[PAWM]
PROPOSED ACTION (the action you are currently about to take):
@@{proposed next action}@@

PITFALL ANTICIPATION (an objective preview from a world model that rolled your PROPOSED ACTION forward over the next several steps - a forecast of what could go wrong):
@@{world model output}@@

This PITFALL ANTICIPATION means your PROPOSED ACTION may lead into a pitfall. Consider this forecast, then decide using the current state above: if it changes your mind, pick a different action; otherwise keep your PROPOSED ACTION and proceed. Respond in the required format with exactly one action.
\end{promptbox}

\subsection{WebArena}
Similar to ScienceWorld, here we also list the system prompt, user prompt, agent history format and action reproposal prompt for WebArena Agent. In WebArena, the observation at each step is the current webpage. We parse its raw HTML into a simplified format similar to an accessibility tree but keeping the semantic ID of each element, so that the agent can refer to elements directly in its actions. Since one single webpage is already long in context, the agent is required to summarize the current page and lay out its plan before producing an action, and it is this summary and plan, rather than the raw page, that is stored in the agent history as the compressed observation.

\begin{promptbox}{Agent History}
You are an autonomous intelligent agent tasked with navigating a web browser to complete specific objectives. You have access to function calling through the step_browser tool to interact with web pages.

ALLOWED SITES:
@@{allowed_sites}@@
Do NOT navigate to, search on, or query any external website (e.g., google.com, bing.com, wikipedia.org, maps.google.com, youtube.com, github.com, etc.).

AVAILABLE TOOL:
**step_browser** - Execute browser actions and get observations

BROWSER ACTIONS (Function Call Parameters):
The step_browser tool accepts action objects with the following available actions:

1. CLICK ELEMENT - Click on any clickable element
   {"action": "click", "target": "semantic_id"}
   - target: The semantic ID of the element to click (REQUIRED)

2. TYPE TEXT - Type text into input fields with optional Enter key press
   {"action": "type", "target": "semantic_id", "text": "content", "enter": true}
   - target: The semantic ID of the input element (REQUIRED)
   - text: The text content to type (REQUIRED)
   - enter: Whether to press Enter after typing (OPTIONAL, default: false)

3. HOVER ELEMENT - Hover over an element to trigger tooltips or dropdowns
   {"action": "hover", "target": "semantic_id"}
   - target: The semantic ID of the element to hover over (REQUIRED)

4. SELECT OPTION - Select an option from dropdown/select elements
   {"action": "select", "target": "semantic_id", "value": "option_value"}
   - target: The semantic ID of the select element (REQUIRED)
   - value: The value of the option to select (REQUIRED)

5. CLEAR INPUT - Clear the content of an input element
   {"action": "clear", "target": "semantic_id"}
   - target: The semantic ID of the input element to clear (REQUIRED)

6. KEY PRESS - Press keyboard keys, optionally on specific elements
   {"action": "key_press", "key": "Enter", "target": "semantic_id"}
   - key: The key to press (e.g., "Enter", "Escape", "Tab", "ArrowDown") (REQUIRED)
   - target: The semantic ID of element to focus before key press (OPTIONAL)

7. NAVIGATE TO URL - Navigate to a specific URL in current tab
   {"action": "goto_url", "url": "https://example.com"}
   - url: The URL to navigate to (REQUIRED)

8. GO BACK - Navigate back in browser history
   {"action": "back"}
   - No parameters required

9. GO FORWARD - Navigate forward in browser history
   {"action": "forward"}
   - No parameters required

10. REFRESH PAGE - Reload the current page
    {"action": "refresh"}
    - No parameters required

11. NEW TAB - Open a new browser tab
    {"action": "new_tab", "url": "https://example.com"}
    - url: URL to open in the new tab (OPTIONAL, opens blank tab if not provided)

12. SWITCH TAB - Switch to a different browser tab
    {"action": "switch_tab", "tab_id": 0}
    - tab_id: The ID number of the tab to switch to (REQUIRED, starts from 0)

13. CLOSE TAB - Close a specific browser tab
    {"action": "close_tab", "tab_id": 0}
    - tab_id: The ID number of the tab to close (REQUIRED, starts from 0)

14. TERMINATE TASK - End the task with final answer
    {"action": "terminate", "answer": "final answer or result"}
    - answer: Your final answer or result for the task (OPTIONAL, use empty string if no specific answer)

READING THE OBSERVATION:
The current page is given as a SIMPLIFIED ACCESSIBILITY TREE: an indented list of elements, one per line, in the form
    - role [id] "name" [state]
    - `role` is what the element is (link, button, textbox, checkbox, radio, combobox, menuitem, heading, row, cell, list, ...).
    - `"name"` is the element's visible label/text.
    - Indentation shows containment (e.g. rows and cells nested inside a table, items inside a list).
    - `[id]` is the element's semantic ID. Use this exact id as the `target` of click/type/hover/select/clear/key_press actions. ONLY elements that show an `[id]` can be acted on; lines without an `[id]` are static text/structure for context.
    - A link's destination, when meaningful, appears on the next indented line as `- /url: ...`.
    - For textboxes, comboboxes, and selects, the current value is shown after a colon, e.g. `combobox [period] "Period": month`.
    - Bracketed states describe status: `[focused]`, `[disabled]`, `[readonly]`, `[expanded=true]`, `[selected]`, `[checked=true]`.
    - For menu items, try hovering first to see if dropdown options appear.
    The URL and title of the current page are shown above the tree, and the open browser tabs are listed below it.

STRATEGY AND GUIDELINES:
1. **Start with step_browser** - Take an initial action to see the current page state
2. **Summarize the observation** - On EVERY step, before doing anything else, output a line prefixed with "SUMMARIZATION:" that briefly summarizes the main content of the current page's accessibility tree (what kind of page it is and its key elements/sections), and explicitly calls out the parts most relevant to the OBJECTIVE (the links, inputs, results, or data needed to make progress)
3. **Plan the next step** - After the summary, output a line prefixed with "PLANNING:" that, based on that summary, decides the single next action that makes the most progress toward the objective
4. **Execute actions** - Only after writing the SUMMARIZATION and PLANNING lines, use the step_browser tool to take that action
5. **Verify progress** - Check each observation to confirm actions had the expected effect
6. **Handle failures** - If an action fails, try alternative approaches or element targets
7. **Complete appropriately** - Use terminate action only when the task is fully complete

TERMINATION RULES:
- If the task is a QUESTION ANSWERING task (e.g., "What is the price of X?", "What are the features of Y?"), use the terminate action with your answer once you find the information
- For ALL OTHER TASKS (purchases, form filling, navigation, etc.), NEVER use the terminate action until the task is completely finished

EXAMPLES:
- Click a button: {"action": "click", "target": "login_button"}
- Search with Enter: {"action": "type", "target": "search_box", "text": "meat substitutes", "enter": true}
- Select dropdown: {"action": "select", "target": "category_dropdown", "value": "food"}
- Complete task: {"action": "terminate", "answer": "Product XYZ costs \$150"}

TOOL CALL EXAMPLES:
Click action:
step_browser(action="click", target="add_to_cart_btn")

Terminate action with answer:
step_browser(action="terminate", answer="The Samsung Galaxy headphones cost \$89.99")

Start by taking an action to see the current page state, then work systematically toward completing the objective.

You should ALWAYS RETURN A SINGLE TOOL CALL and make another tool call after you get the updated page state.
When you believe you have completed the task, you should use the terminate tool call.
\end{promptbox}

\begin{promptbox}{User Prompt}
OBJECTIVE:
@@{task instruction}@@

AGENT HISTORY (prior thinking and actions, web pages omitted for brevity):
@@{agent history}@@

CURRENT BROWSER STATE:
@@{current webpage}@@

Decide the next action and perform exactly one step_browser tool call.
\end{promptbox}

\begin{promptbox}{Agent History}
--- Step 1 ---
SUMMARIZATION: @@{step 1 webpage summary}@@
PLANNING: @@{step 1 planning}@@
ACTION: @@{step 1 action}@@

--- Step 2 ---
SUMMARIZATION: @@{step 2 webpage summary}@@
PLANNING: @@{step 2 planning}@@
ACTION: @@{step 2 action}@@

...
\end{promptbox}

\begin{promptbox}{Action Reproposal Prompt after User Prompt}
[Next-Obv]
PROPOSED ACTION (the action you are currently about to take):
@@{proposed next action}@@

PREDICTED NEXT OBSERVATION (a world model rolled your PROPOSED ACTION forward one step and predicts the page you would then be looking at - a forecast of the resulting state, not an instruction):
@@{world model output}@@

Use this PREDICTED NEXT OBSERVATION to sanity-check your PROPOSED ACTION: judge for yourself, using the CURRENT BROWSER STATE, whether landing on that predicted page is the progress you intend. If the prediction shows your action would fail, stall, or lead somewhere unproductive, pick a better action; if the predicted next state looks like real progress, keep your PROPOSED ACTION and proceed. 

[Full-Traj]
PROPOSED ACTION (the action you are currently about to take):
@@{proposed next action}@@

ANTICIPATION (a forecast from a world model that rolled your PROPOSED ACTION forward to the end of the task):
@@{world model output}@@

Use this ANTICIPATION to sanity-check your PROPOSED ACTION: judge for yourself, using the CURRENT BROWSER STATE, whether landing on that predicted page is the progress you intend. If the anticipation shows your action would fail, stall, or lead somewhere unproductive, pick a better action; if the predicted next state looks like real progress, keep your PROPOSED ACTION and proceed. 

[SubGoal-Traj]
PROPOSED ACTION (the action you are currently about to take):
@@{proposed next action}@@

ANTICIPATION (a forecast from a world model that rolled your PROPOSED ACTION forward to the next milestone):
@@{world model output}@@

Use this ANTICIPATION to sanity-check your PROPOSED ACTION: judge for yourself, using the CURRENT BROWSER STATE, whether landing on that predicted page is the progress you intend. If the anticipation shows your action would fail, stall, or lead somewhere unproductive, pick a better action; if the predicted next state looks like real progress, keep your PROPOSED ACTION and proceed. 

[PAWM]
PROPOSED ACTION (the action you are currently about to take):
@@{proposed next action}@@

PITFALL ANTICIPATION (an objective preview from a world model that rolled your PROPOSED ACTION forward over the next several steps - a forecast of what could go wrong)::
@@{world model output}@@

This PITFALL ANTICIPATION means your PROPOSED ACTION may lead into a pitfall. Consider this forecast, then decide using the CURRENT BROWSER STATE above: if it changes your mind, pick a different action; otherwise keep your PROPOSED ACTION and proceed. 
\end{promptbox}

\section{Experiment Details of Pilot Study}
This section provides additional experiment details in our pilot study.

\subsection{SFT Details for World Models}
\label{app:sft}
Here we list the SFT details for training the world models with varied output contents in Section~\ref{sec:what}.

\noindent \textbf{SFT Dataset Construction.} To collect the data for training these world models, we first sample trajectories on the training split of both datasets by running the baseline setting (N/A) with Claude Sonnet 4.5 as the agent, repeating each task three times at a generation temperature of $1.0$. This results in 2,124 agent trajectories for ScienceWorld and 1,941 for WebArena. Since a world model is meant to reflect the environment's response to each action, failed trajectories remain useful, as they show how a poor action leads to an unfavorable observation. We therefore split each collected trajectory at every time step $t$ and prompt Sonnet 4.5 to summarize the content that follows as the world model's target output. The prompts used to generate targets for each setting are listed below.

The agent future field follows the same format as the agent history, except that it covers the steps taken after the current one. Placeholders with multiple variants are listed at the end of each prompt box, where the tag in square brackets marks the condition and the text following it is the content substituted into the placeholder under that condition. For example, {outcome} is replaced with the [SUCCESS] line when the collected trajectory succeeded and with the [FAILURE] line when it failed. For Next-Obv on WebArena-Lite, no additional summarization is needed, as the agent already summarizes each page it sees as part of its output, so the summary produced at step $t+1$ serves directly as the target next observation.

\begin{promptbox}{System Prompt for Next-Obv WM Targets Generation on ScienceWorld}
You are labeling training data for a ScienceWorld NEXT-OBSERVATION world model. A separate "actor" agent plays the game; at some step it PROPOSED an action, and the game then transitioned to a new state. Your job is to rewrite that ACTUAL RESULTING STATE into a concise, natural-language PREDICTION of what the actor will be looking at right after the action executes - the target a world model should learn to produce from the state and the proposed action ALONE.

You are given the TASK, the AGENT HISTORY, the CURRENT STATE the actor saw before acting, the PROPOSED NEXT ACTION, and - as ground truth for labeling only - the ACTUAL RESULTING STATE the game transitioned to. That resulting state has THREE parts, all after the action executed: the RESULT OBSERVATION (the game's text response), the RESULTING LOCATION (the room the actor is now in and what it contains), and the RESULTING INVENTORY (what the actor is now carrying). Summarize the WHOLE resulting state - not just the terse result line - into the world model's spoken prediction.

ABOUT SCIENCEWORLD (so your summary is faithful):
- It is a text simulation of elementary-school science experiments. The actor moves between rooms and manipulates objects to complete a scientific task. Common actions: move/teleport, open/close, pick up / put down / move objects, activate/deactivate devices (stove, freezer, blast furnace, sink), use a thermometer, mix, read, wait (let time pass so heating/cooling/growth happens), and "focus on X" (a one-time, irreversible commitment marking the task target).
- The resulting state is the combination of: the game's terse text response (RESULT OBSERVATION), plus how the world now looks (RESULTING LOCATION - the current room, its objects, open/closed doors, devices on/off) and what is held (RESULTING INVENTORY). A single action can change any of these: entering a new room, an object moving into/out of the inventory, a device turning on, a thermometer reading, a container's contents, a door opening, "focus" ending the episode, or "No known action"/no change.

HOW TO WRITE THE PREDICTION:
- Describe the SINGLE resulting state - one step from now - not the end of the task. Do not forecast multiple steps ahead.
- Cover the WHOLE state transition, drawing on all three parts: what the game reported (result observation), WHERE the actor now is if it changed (resulting location), and WHAT it is now carrying if it changed (resulting inventory). Weave them into one coherent prediction rather than listing them as separate fields.
- Be FAITHFUL to the ACTUAL RESULTING STATE: report what actually happened, including a "bad" outcome (no change, an invalid action, a mislabeled target, a device that did nothing because a precondition was unmet, or "focus" ending the episode). Never sugar-coat it into the outcome the actor hoped for - a truthful prediction of a bad next state is exactly the signal the world model must learn.
- Preserve TASK-RELEVANT DETAILS verbatim: measurement readings and units (e.g. "35 degrees"), exact object/substance names, on/off and open/closed states, which room the actor moved to, what entered or left the inventory, and whether the episode ended. Keep numbers and names character-exact.
- FOCUS ON WHAT MATTERS FOR THE TASK: summarize the parts of the resulting state relevant to making progress (or the failure), and drop irrelevant scenery. A couple of sentences is ideal; do not pad, and do not restate the entire room object-list unless it changed in a way that matters to the task.
- Write it as a PREDICTION in the second person ("You ..."), the voice a world model uses when telling the actor where its action lands. Do NOT mention that you were shown the actual result, and do NOT reference "ground truth" or "the label".
- Do NOT output literal actions, object IDs, or code. Plain natural language only.

OUTPUT FORMAT:
- Output ONLY the predicted next observation, wrapped in <anticipation> and </anticipation> tags. Nothing else. For example:
  <anticipation>You are now in the kitchen holding the metal pot, but it is still empty: the sink was never turned on, so pouring did nothing and nothing entered the pot. The room is unchanged and you would need to activate the sink before any water is available.</anticipation>
\end{promptbox}

\begin{promptbox}{User Prompt for Next-Obv WM Targets Generation on ScienceWorld}
TASK:
@@{task instruction}@@

AGENT HISTORY (prior thoughts, actions, and short observations):
@@{agent history}@@

CURRENT STATE (room, inventory, last observation - what the actor saw BEFORE acting):
@@{current state}@@

PROPOSED NEXT ACTION (the actor executed this):
@@{proposed next action}@@

ACTUAL RESULTING STATE (ground truth - the whole state AFTER the action; label ONLY, the world model does not see this):
  RESULT OBSERVATION (the game's response to the action):
@@{result observation}@@
  RESULTING LOCATION (the room after the action):
@@{result location}@@ or '(unchanged/unknown)'
  RESULTING INVENTORY (what you are carrying after the action):
@@{result inventory}@@ or '(unchanged/unknown)'

Summarize this WHOLE ACTUAL RESULTING STATE - the result observation together with the resulting location and inventory - into the concise, task-focused, second-person PREDICTED NEXT OBSERVATION as specified. Output ONLY the <anticipation>...</anticipation> block.
\end{promptbox}

\begin{promptbox}{System Prompt for Full-Traj \& SubGoal-Traj WM Targets Generation on ScienceWorld}
You are labeling training data for a ScienceWorld TRAJECTORY world model. A separate "actor" agent plays the game; at some step it PROPOSED an action, and the run then went on to unfold in a particular way. Your job is to rewrite the AGENT FUTURE - the steps the run ACTUALLY took from that point - into a concise, forward-looking PREDICTION of the trajectory that the proposed action sets in motion.

You are given the TASK, the AGENT HISTORY (the actor's prior steps and the state it currently faces), the AGENT FUTURE (ground truth - the actions the run actually took from here and the states they produced), and the RUN OUTCOME (whether the run ultimately SUCCEEDED or FAILED). Turn that ground-truth future into the world model's spoken forecast - the target it should learn to produce from the state and the proposed action ALONE.

ABOUT SCIENCEWORLD (so your forecast is faithful):
- It is a text simulation of elementary-school science experiments. The actor moves between rooms and manipulates objects to complete a scientific task. Common actions: move/teleport, open/close, pick up / put down / move objects, activate/deactivate devices (stove, freezer, blast furnace, sink), use a thermometer, mix, read, wait (let time pass so heating/cooling/growth happens), and "focus on X" (a one-time, irreversible commitment marking the task target).
- State changes take TIME and PRECONDITIONS: heating/cooling require the substance on an active device and repeated waiting; containers must be open; a substance must be identified and in hand before focusing. `focus on` the wrong object, or before the correct target is ready, usually FAILS the task with a large penalty and ends the episode.

HOW TO WRITE THE FORECAST:
- Trace the ARC of the trajectory the AGENT FUTURE shows - the sequence of actions and the states they produce - as a coherent path, NOT a per-step transcript. A few sentences that convey where the path goes and how it ends (within the horizon you are told to stop at).
- Be FAITHFUL TO THE RUN OUTCOME. If the run SUCCEEDED, forecast the path reaching the goal. If it FAILED, forecast the failure HONESTLY - the wrong focus, the wasted loop, the missing precondition, the stall, the wrong/early termination. NEVER pretend a failing run succeeds; a truthful forecast of a bad trajectory is exactly what lets the actor reconsider.
- Ground it in what the AGENT FUTURE actually contains. Do not invent steps that did not happen. Predictions grow less certain further out, so keep the later stretch appropriately tentative in wording, but stay faithful to the recorded future.
- Preserve TASK-RELEVANT DETAILS: the correct/incorrect target focused on, key state changes (a device turned on, a substance reaching a temperature or changing state), and whether the episode ends. Keep object/substance names and numeric readings character-exact when they matter.
- Write it as a PREDICTION in the second person ("You ..."), the voice a world model uses when telling the actor where its action leads. Do NOT mention that you were shown the actual future, the run outcome label, or "ground truth".
- Do NOT output literal actions, object IDs, or code, and do NOT number the steps. Plain natural-language prose.

OUTPUT FORMAT:
- Output ONLY the forecast, wrapped in <summary> and </summary> tags. Nothing else. For example:
  <summary>You move the metal pot with the water onto the stove, turn the stove on, and wait repeatedly as the temperature climbs; the water reaches its boiling point and changes to steam, and once you focus on the correct substance the task completes successfully.</summary>
\end{promptbox}

\begin{promptbox}{User Prompt for Full-Traj \& SubGoal-Traj WM Targets Generation on ScienceWorld}
TASK:
@@{task instruction}@@

AGENT HISTORY (the actor's prior steps and the state it currently faces):
@@{agent history}@@

CURRENT STATE (room, inventory, last observation - what the actor sees now):
@@{current state}@@

AGENT FUTURE (starting from the actor's decision at the current step, the steps this run ACTUALLY took - ground truth, not a guess):
@@{agent future}@@

RUN OUTCOME: this trajectory ultimately @@{outcome}@@.

@@{summary instruction}@@ Summarize it as a coherent forward-looking anticipation in the second person, faithful to the RUN OUTCOME (if it failed, forecast the failure honestly; do not pretend it succeeds). Wrap your final summary inside <summary> and </summary> tags.

================================================
NOTE (not part of the prompt): the placeholders below are each instantiated as one of the listed variants.
================================================

@@{outcome}@@
  [SUCCESS] SUCCEEDED - it reached the goal and the task was scored complete
  [FAILURE] FAILED - it did not complete the task correctly

@@{summary instruction}@@
  [Full-Traj] Summarize the AGENT FUTURE all the way to how the task TERMINATES (the actor's final action and whether it ends in success or failure).
  [SubGoal-Traj] Summarize the AGENT FUTURE only UP TO THE NEXT MILESTONE - the next key step of progress toward the task goal (reaching the right room, getting the correct substance/object in hand, a device turned on with the target inside, the substance reaching the required temperature or state, or focusing on the correct target). Stop once that milestone is reached, or once the attempt to reach it clearly fails; do not summarize past it. Do not name the milestone explicitly - just describe the path up to that stopping point.
\end{promptbox}

\begin{promptbox}{System Prompt for Full-Traj \& SubGoal-Traj WM Targets Generation on WebArena}
You are an expert annotator who writes world-model summaries of how a web-navigation task unfolds. A separate "actor" agent drives the browser; your job is to summarize its future, not to take actions yourself.

ABOUT THE ACTOR (a black box to you):
- The actor is an autonomous web agent that completes the objective by calling a single tool, `step_browser`, once per turn. Each call performs one browser action and returns the resulting page state.
- `step_browser` can: click an element, type text into a field (optionally pressing Enter), hover, select a dropdown value, clear a field, press a key, navigate to a URL, go back/forward, refresh, open/switch/close a browser tab, and finally terminate the task while reporting an answer.
- Each turn the actor records an OBSERVATION SUMMARIZATION (what it observed that turn), a PLANNING (its reasoning), and the ACTION it then took. One turn's action produces the next turn's observation. One step = one action together with the page it produces.

WHAT YOU ARE GIVEN:
- The TASK OBJECTIVE.
- The AGENT HISTORY: the steps the actor has already taken up to the current state.
- The AGENT FUTURE: the remaining steps this run ACTUALLY took (GROUND TRUTH, not a guess), possibly truncated to a horizon you are told about below. Each future step shows the observation the actor saw, its planning, and the action it took.
- The RUN OUTCOME: whether this run ultimately SUCCEEDED or FAILED. This is the real outcome of the trajectory you are summarizing.

YOUR JOB:
Write a single coherent, forward-looking description of what happens from the current state onward, as a STEP-BY-STEP CHAIN that follows the AGENT FUTURE exactly. The OUTPUT SHAPE is fixed:
- START with the OBSERVATION the PROPOSED NEXT ACTION produces - the page state that results right after that action executes.
- THEN, for each subsequent step in the AGENT FUTURE, in order: the ACTION the actor takes and the OBSERVATION that action produces (action -> resulting page state), chained one step after another.
This reads as: proposed-action's result, then next action and its result, then the next, and so on - one clause per step, following the real AGENT FUTURE turn by turn.

Rules:
- FOLLOW THE GROUND TRUTH, STEP BY STEP: describe the ACTUAL steps in the AGENT FUTURE, in their real order. Do NOT skip, merge, or reorder steps - including routine navigation (clicking into a menu/section, opening a form, focusing a field); each of these is its own step and gets its own clause. Do NOT jump ahead to a later step's outcome, and do NOT invent steps that are not in the AGENT FUTURE.
- FAITHFUL TO THE OUTCOME: describe the ACTUAL path, not an idealized one. If the run SUCCEEDED, follow the route to the correct result and the actor terminating with the answer. If the run FAILED, describe honestly where it goes wrong - the action that errors, the empty/wrong result it lands on, the loop it gets stuck in, or its terminating with an unverified/incorrect answer. Do NOT pretend a failing run succeeds, and do NOT invent a success it never reached.
- NO FUTURE-SPECIFIC VALUES: the summary is an ANTICIPATION of an unseen future, so it must not leak concrete values the current step could not yet know - observed result names, counts, prices, dates, IDs, usernames, or the final answer. Refer to these by role and location ("you open the customer's profile", "the order total shown here is the answer"), never by literal value. A value may be stated verbatim ONLY if it is already known now, i.e. it comes from the TASK OBJECTIVE or AGENT HISTORY. A value the actor TYPES is not automatically allowed: if it was discovered earlier in the future, it is still future-specific ("you search for the customer you found", not the literal name). Leaking such values trains the reader to hallucinate the contents of pages it has not seen.

END CRITERION (WHERE TO STOP):
- The ask states WHERE the chain ends - to how the task TERMINATES, after the next ~5 steps, or up to the next milestone. Follow the AGENT FUTURE step by step until that end point and STOP there. Do not continue past it, and do not forecast steps beyond what the AGENT FUTURE shows.

IMPORTANT:
- Address the actor in the second person ("You click ...", "the page now shows ...").
- Describe both the actions and the resulting page states in plain natural language. Do NOT output literal browser actions, tool calls, JSON, semantic IDs, or code.

OUTPUT FORMAT:
- Wrap your final summary in <summary> and </summary> tags, e.g. <summary>You click into the Reports menu and it expands to list the report categories. You choose Bestsellers, and the page loads a report view with a filter form above a product-ranking table. You set the period dropdown to the year the objective asks about, and the form updates to expose the date-range fields. You click the From date field and it becomes focused. You type the start date into it, and the field now shows the entered date.</summary>
\end{promptbox}

\begin{promptbox}{User Prompt for Full-Traj \& SubGoal-Traj Targets Generation on WebArena}
TASK OBJECTIVE:
@@{task instruction}@@

AGENT HISTORY (the actor's prior steps and the observation it currently faces, web pages omitted for brevity):
@@{agent history}@@

AGENT FUTURE (starting from the actor's decision at the current step, the steps this run actually took - ground truth, not a guess):
@@{agent future}@@

RUN OUTCOME: this trajectory ultimately @@{outcome}@@.

@@{summary instruction}@@ Summarize it as a coherent forward-looking anticipation in the second person, faithful to the RUN OUTCOME (if it failed, forecast the failure honestly; do not pretend it succeeds). Wrap your final summary inside <summary> and </summary> tags.

================================================
NOTE (not part of the prompt): the placeholders below are each instantiated as one of the listed variants.
================================================

@@{outcome}@@
  [SUCCESS] SUCCEEDED - it reached and reported the correct result
  [FAILURE] FAILED - it did not complete the task correctly

@@{summary instruction}@@
  [Full-Traj] END CRITERION: continue the step-by-step sequence all the way to how the task TERMINATES (the actor's final action and whether it ends in success or failure).
  [SubGoal-Traj] END CRITERION: continue the step-by-step sequence only UP TO THE NEXT MILESTONE - the next key step of progress toward the objective (reaching the right section/report, a filter's results populating, opening the target entity, a form submission confirmed, or reading the value that answers the objective) - then stop, or stop once the attempt to reach it clearly fails. Do not name the milestone explicitly.
\end{promptbox}

Given the generated targets, we also need the prompts that serve as the world model's input to form the final training dataset. The world model input prompts for each setting, which are also used at inference time, are presented below.

\begin{promptbox}{System Prompt for Next-Obv WM on ScienceWorld}
You are a ScienceWorld world model. A separate "actor" agent actually plays the game; it has ALREADY chosen the single action it is about to take. Your job is a plain world-model prediction: given the CURRENT STATE and the PROPOSED NEXT ACTION, predict the NEXT OBSERVATION - the game's textual response and the resulting state the actor will see right after that action executes. You do this on EVERY turn, for EVERY proposed action, whether it is good, bad, or routine. You are a predictive model of the environment, NOT a supervisor: you forecast what the game will report, you never command the actor or take actions yourself. The actor, which can see the full room, decides what to do with your prediction.

ABOUT THE ENVIRONMENT AND ACTOR (a black box to you):
- ScienceWorld is a text simulation of elementary-school science experiments. The actor controls a character in a house of rooms (kitchen, bathroom, workshop, greenhouse, foundry, etc.) containing objects it can inspect and manipulate to complete a scientific task.
- Each turn the actor takes ONE action from a menu of currently-valid actions, e.g.: move between rooms (teleport/go/open door), pick up / put down / move objects, activate/deactivate devices (stove, freezer, blast furnace, sink), use a thermometer, mix chemicals, read, wait (let time pass so heating/cooling/growth happens), and "focus on X".
- One mechanic worth knowing: `focus on X` is a one-time, task-defining commitment that marks X as the target object; it is irreversible for the episode, and focusing on the wrong object or focusing prematurely usually ends the episode with a large penalty.
- You do NOT see the raw environment internals. You see the TASK, the actor's AGENT HISTORY (per prior turn: its thought, the ACTION it took, and a short observation), the CURRENT STATE (current room description + inventory + last observation), and - most importantly - the PROPOSED NEXT ACTION the actor is about to execute.

YOUR JOB - PREDICT THE NEXT OBSERVATION:
- Roll the environment forward exactly ONE step from the CURRENT STATE, applying the PROPOSED NEXT ACTION, and describe the observation the game would return: the text response and how the relevant state changes (room contents, inventory, a device turning on, a thermometer reading, a container's contents, a door opening, an object being moved, "focus" ending the episode, or "No known action"/no change).
- Predict the ACTUAL outcome, not the desirable one. If the action would produce no real state change, target a mislabeled object (e.g. a "tin cup" is a container made of tin, not the substance "tin"), operate a device whose precondition is not met (container closed, device off, nothing inside), or be rejected as invalid, predict THAT outcome faithfully - do not sugar-coat it. A truthful prediction of a bad next state is exactly what helps the actor reconsider.
- Ground the prediction in the CURRENT STATE, ScienceWorld's mechanics (heating/cooling take repeated "wait" steps; containers must be open; devices must be activated; a substance must be identified and in hand before focusing), and what the AGENT HISTORY reveals about how this episode is unfolding.
- Keep it to the IMMEDIATE next observation - the single step from now - not the end of the task.

IMPORTANT:
- Do NOT output literal actions or object IDs. Describe the predicted observation in plain natural language.
- Keep it brief - a few short sentences focused on what the next observation will be.
- Stay neutral and descriptive: you are forecasting the next observation for the actor's consideration, not instructing it.

OUTPUT FORMAT:
- Always output your predicted next observation wrapped in <anticipation> and </anticipation> tags, e.g.
  <anticipation>The action returns no useful change: the sink has not been activated, so the cup comes back still empty and the actor is looking at the same room with an empty cup in its inventory. It would still need to turn on the sink before water is available.</anticipation>
\end{promptbox}

\begin{promptbox}{User Prompt for Next-Obv WM on ScienceWorld}
TASK:
@@{task instruction}@@

AGENT HISTORY (prior thoughts, actions, and short observations):
@@{agent history}@@

CURRENT STATE (room, inventory, last observation):
@@{current state}@@

PROPOSED NEXT ACTION (the actor is about to execute this):
@@{proposed next action}@@

Predict the NEXT OBSERVATION - the state the actor will be looking at right after this action executes. Roll the environment forward exactly one step and describe the resulting observation (the game's text response and how the state changes), predicting the ACTUAL outcome whether it advances the task or fails. Wrap your predicted next observation inside <anticipation> and </anticipation> tags.
\end{promptbox}

\begin{promptbox}{System Prompt for Full-Traj WM on ScienceWorld}
You are an expert ScienceWorld trajectory world model. A separate "actor" agent actually plays the game; it has ALREADY chosen the single action it is about to take. Your job is to FORECAST the full future trajectory that action sets in motion - all the way to how the task would end. You are a predictive model of the environment, NOT a supervisor: you describe what would plausibly happen, you never command the actor or take actions yourself. The actor, which can see the full room, decides what to do with your forecast.

ABOUT THE ENVIRONMENT AND ACTOR (a black box to you):
- ScienceWorld is a text simulation of elementary-school science experiments. The actor controls a character in a house of rooms (kitchen, bathroom, workshop, greenhouse, foundry, etc.) containing objects it can inspect and manipulate to complete a scientific task.
- Each turn the actor takes ONE action from a menu of currently-valid actions, e.g.: move between rooms (teleport/go/open door), pick up / put down / move objects, activate/deactivate devices (stove, freezer, blast furnace, sink), use a thermometer, mix chemicals, read, wait (let time pass so heating/cooling/growth happens), and "focus on X".
- CRITICAL game mechanic - `focus on X`: a one-time, task-defining commitment marking X as the target; focusing on the WRONG object, or before the correct target is identified and in hand, usually FAILS the task immediately with a large negative score and ends the episode. Focusing is irreversible for the episode.
- You do NOT see the raw environment internals. You see the TASK, the actor's AGENT HISTORY (per prior turn: its thought, the ACTION it took, and a short observation), the CURRENT STATE (current room description + inventory + last observation), and - most importantly - the PROPOSED NEXT ACTION the actor is about to execute.

FORECAST THE FUTURE TRAJECTORY (the anticipation):
- For the proposed action, roll the environment forward from it and predict the whole TRAJECTORY the actor is on: the sequence of actions it would go on to take and the state each one produces, continuing ALL THE WAY to how the task would terminate - either SUCCEEDING (reaching and focusing on / achieving the correct outcome, scored 100) or FAILING (focusing on the wrong target, erroring out, stalling, looping without progress, running out of steps, or ending before the required state change happened). Predict the ACTUAL ending you expect, good or bad - do NOT assume it ends in success.
- This is a plain world-model prediction of the entire path forward, NOT limited to problems. Ground it in the CURRENT STATE, ScienceWorld's mechanics (heating/cooling take repeated "wait" steps; containers must be open; devices must be activated; a substance must be identified and in hand before focusing), and what the AGENT HISTORY reveals. Predictions grow less certain further out, so keep the later stretch appropriately tentative.
- Keep it compact - a few short sentences tracing the arc of the path to its end, not a per-step script. Do NOT output literal actions or object IDs; describe the actions and resulting states in plain language.

OUTPUT FORMAT:
- Output ONLY the forecast trajectory, wrapped in <anticipation> and </anticipation> tags. Nothing else.
- Example (trajectory that fails):
  <anticipation>Focusing on the tin cup now would commit the target to a container made of tin rather than the substance "tin" the task wants, ending the episode immediately with a large penalty - the trajectory terminates in failure without ever heating the real target.</anticipation>
- Example (trajectory that succeeds):
  <anticipation>Moving the metal pot onto the stove sets up the heat source; the actor then activates the stove and waits repeatedly as the substance climbs to its boiling point, focuses on the correct substance once it is boiling, and the episode ends scored 100 - a clean path to success.</anticipation>
\end{promptbox}

\begin{promptbox}{System Prompt for SubGoal-Traj WM on ScienceWorld}
You are an expert ScienceWorld trajectory world model. A separate "actor" agent actually plays the game; it has ALREADY chosen the single action it is about to take. Your job is to FORECAST the trajectory that action sets in motion up to the NEXT MILESTONE. You are a predictive model of the environment, NOT a supervisor: you describe what would plausibly happen, you never command the actor or take actions yourself. The actor, which can see the full room, decides what to do with your forecast.

ABOUT THE ENVIRONMENT AND ACTOR (a black box to you):
- ScienceWorld is a text simulation of elementary-school science experiments. The actor controls a character in a house of rooms (kitchen, bathroom, workshop, greenhouse, foundry, etc.) containing objects it can inspect and manipulate to complete a scientific task.
- Each turn the actor takes ONE action from a menu of currently-valid actions, e.g.: move between rooms (teleport/go/open door), pick up / put down / move objects, activate/deactivate devices (stove, freezer, blast furnace, sink), use a thermometer, mix chemicals, read, wait (let time pass so heating/cooling/growth happens), and "focus on X".
- CRITICAL game mechanic - `focus on X`: a one-time, task-defining commitment marking X as the target; focusing on the WRONG object, or before the correct target is identified and in hand, usually FAILS the task immediately with a large negative score and ends the episode. Focusing is irreversible for the episode.
- You do NOT see the raw environment internals. You see the TASK, the actor's AGENT HISTORY (per prior turn: its thought, the ACTION it took, and a short observation), the CURRENT STATE (current room description + inventory + last observation), and - most importantly - the PROPOSED NEXT ACTION the actor is about to execute.

WHAT A MILESTONE IS:
- A milestone is the next meaningful step of progress on the way to the task goal - a point where a distinct sub-result is achieved. Examples: reaching the right room, getting the correct substance/object into hand, moving a container onto the heat/cold source, the substance reaching the target temperature or state, or focusing on the correct target. It is the natural "am I on track?" boundary, not a fixed number of steps away - it may be one step or several.

FORECAST TO THE NEXT MILESTONE (the anticipation):
- For the proposed action, figure out (for yourself) the next milestone the actor is working toward from the CURRENT STATE, and use it only as the STOPPING POINT for your forecast - you do NOT need to state the milestone explicitly. Roll the environment forward from the proposed action and predict the trajectory - the actions it would take and the state each produces - UP TO the moment that milestone is reached (or up to the moment the attempt to reach it clearly fails). Stop there; do not forecast past it. If the milestone is task completion, forecast through termination and say whether it ends in success or failure.
- Predict the ACTUAL path, good or bad - do NOT assume the milestone is reached cleanly. If the action makes solid progress toward it, forecast that; if it would fail to reach it (no change, invalid action, wrong focus, wrong turn, looping), forecast that instead.
- This is a milestone-bounded world-model prediction, NOT limited to problems. Ground it in the CURRENT STATE, ScienceWorld's mechanics (heating/cooling take repeated "wait" steps; containers must be open; devices must be activated; a substance must be identified and in hand before focusing), and what the AGENT HISTORY reveals.
- Keep it brief - a short sketch of the path up to that stopping point, not a per-step script. Do NOT output literal actions or object IDs; describe the actions and resulting states in plain language.

OUTPUT FORMAT:
- Output ONLY the forecast (the path up to the stopping point), wrapped in <anticipation> and </anticipation> tags. Nothing else. Do not preface it by naming the milestone.
- Example (path that fails short of the milestone):
  <anticipation>Pouring into the cup now returns no change because the sink was never activated, so the cup stays empty - the actor would have to go back and turn on the sink before it has any water to work with.</anticipation>
- Example (path that reaches the milestone):
  <anticipation>Moving the pot onto the stove and activating it starts heating; a few waits later the substance reaches its boiling point and the room reports the change of state - the actor is then set up to focus on the correct substance.</anticipation>
\end{promptbox}

\begin{promptbox}{User Prompt for Full-Traj \& SubGoal-Traj WM on ScienceWorld}
TASK:
@@{task instruction}@@

AGENT HISTORY (prior thoughts, actions, and short observations):
@@{agent history}@@

CURRENT STATE (room, inventory, last observation):
@@{current state}@@

PROPOSED NEXT ACTION (the actor is about to execute this):
@@{proposed next action}@@

@@{world model instruction}@@

================================================
NOTE (not part of the prompt): the placeholders below are each instantiated as one of the listed variants.
================================================

@@{world model instruction}@@
  [Full-Traj] ANTICIPATE the future TRAJECTORY that executing the PROPOSED NEXT ACTION sets in motion, all the way to how the task would TERMINATE - the sequence of actions and resulting states likely to unfold, predicted toward the trajectory's ACTUAL ending (the task succeeding, or failing/stalling). Wrap your anticipated trajectory inside <anticipation> and </anticipation> tags.
  [SubGoal-Traj] ANTICIPATE the future TRAJECTORY that executing the PROPOSED NEXT ACTION sets in motion, up to the NEXT MILESTONE (the next key step of progress toward the task goal) - the sequence of actions and resulting states likely to unfold until that milestone is reached, or until the attempt to reach it clearly fails. Wrap your anticipated trajectory inside <anticipation> and </anticipation> tags.
\end{promptbox}

\begin{promptbox}{System Prompt for Next-Obv WM on WebArena}
You are an expert web-navigation world model. A separate "actor" agent actually drives the browser; it has ALREADY chosen the single action it is about to take. Your job is a plain world-model prediction: given the CURRENT BROWSER STATE and the PROPOSED NEXT ACTION, predict the NEXT OBSERVATION - the page state the actor will be looking at right after that action executes. You do this on EVERY turn, for EVERY proposed action, whether the action is good, bad, or routine. You are a predictive model of the environment, NOT a supervisor: you forecast what the page will become, you never command the actor or take actions yourself.

ABOUT THE ACTOR (a black box to you):
- The actor completes the objective by calling a single tool, `step_browser`, once per turn. Each call performs one browser action and returns the resulting page state.
- `step_browser` can: click an element, type text into a field (optionally pressing Enter), hover, select a dropdown value, clear a field, press a key, navigate to a URL, go back/forward, refresh, open/switch/close a browser tab, and finally terminate the task while reporting an answer.
- You do NOT see the true resulting page. You see the actor's AGENT HISTORY (per prior turn: a SUMMARIZATION of what it observed, its PLANNING, and the ACTION it took), the CURRENT BROWSER STATE, and - most importantly - the PROPOSED NEXT ACTION the actor is about to execute.

YOUR JOB - PREDICT THE NEXT OBSERVATION:
- Roll the environment forward ONE step from the CURRENT BROWSER STATE, applying the PROPOSED NEXT ACTION, and describe the resulting page the way a browser state is described: the URL / page the actor will land on, and the key elements or information it should expect to see there (e.g. a specific results table, a filled-in form field, an opened menu, a confirmation banner, an error message, an unchanged page).
- Predict the ACTUAL outcome, not the desirable one. If the action plausibly advances the objective, predict the productive page it produces. If the action is likely to error out, have no effect, target the wrong element, land on an empty/wrong result, or otherwise fail, predict THAT outcome faithfully - do not sugar-coat it. A truthful prediction of a bad next state is exactly what helps the actor reconsider.
- Ground the prediction in the CURRENT BROWSER STATE, the site's mechanics, and what the AGENT HISTORY reveals about how this environment behaves.
- Keep it to the IMMEDIATE next observation - the single page one step from now - not the end of the task.
- If the CURRENT BROWSER STATE already contains the answer or already shows the required side-effect, and the proposed action is to terminate/report it, predict that the task ends in that terminal state.

IMPORTANT:
- Do NOT output literal browser actions, tool calls, semantic IDs, or code. Describe the predicted page state in plain natural language.
- Keep it brief - a few short sentences focused on what the next page will look like.
- Stay neutral and descriptive: you are forecasting the next observation for the actor's consideration, not instructing it. The actor, who can see the real page, decides what to do with your prediction.

OUTPUT FORMAT:
- Always output your predicted next observation wrapped in <anticipation> and </anticipation> tags, e.g.
  <anticipation>The Bestsellers report page reloads but the table comes back empty ("We couldn't find any records"): the Period is still "Day" while the date range spans a month, so the query returns no rows. The actor would be looking at an empty report and still need to fix the Period setting.</anticipation>
\end{promptbox}

\begin{promptbox}{User Prompt for Next-Obv WM on WebArena}
TASK OBJECTIVE:
@@{task instruction}@@

AGENT HISTORY (prior thinking and actions, web pages omitted for brevity):
@@{agent history}@@

CURRENT BROWSER STATE:
@@{current webpage}@@

PROPOSED NEXT ACTION (the actor is about to execute this):
@@{proposed next action}@@

Predict the NEXT OBSERVATION - the page state the actor will be looking at right after this action executes. Roll the environment forward exactly one step and describe the resulting page (its URL/page and the key elements or information visible there), predicting the ACTUAL outcome whether it advances the objective or fails. Wrap your predicted next observation inside <anticipation> and </anticipation> tags."
\end{promptbox}

\begin{promptbox}{System Prompt for Full-Traj WM on WebArena}
You are an expert web-navigation trajectory world model. A separate "actor" agent actually drives the browser; it has ALREADY chosen the single action it is about to take. Your job is to FORECAST the full future trajectory that action sets in motion - all the way to how the task would end. You are a predictive model of the environment, NOT a supervisor: you describe what would plausibly happen, you never command the actor or take actions yourself.

ABOUT THE ACTOR (a black box to you):
- The actor completes the objective by calling a single tool, `step_browser`, once per turn. Each call performs one browser action and returns the resulting page state.
- `step_browser` can: click an element, type text into a field (optionally pressing Enter), hover, select a dropdown value, clear a field, press a key, navigate to a URL, go back/forward, refresh, open/switch/close a browser tab, and finally terminate the task while reporting an answer.
- You do NOT see the web pages. You see the actor's AGENT HISTORY (per prior turn: a SUMMARIZATION of what it observed, its PLANNING, and the ACTION it took), the CURRENT BROWSER STATE, and - most importantly - the PROPOSED NEXT ACTION the actor is about to execute.

FORECAST THE FUTURE TRAJECTORY (the anticipation):
- For the proposed action, roll the environment forward from it and predict the whole TRAJECTORY the actor is on: the sequence of actions it would go on to take and the page state each one produces, continuing ALL THE WAY to how the task would terminate - either SUCCEEDING (reaching and reporting the correct outcome) or FAILING (erroring out, stalling, looping without progress, or terminating with a wrong/unverified answer). Predict the ACTUAL ending you expect, good or bad - do NOT assume it ends in success.
- This is a plain world-model prediction of the entire path forward, NOT limited to problems. Ground it in the CURRENT BROWSER STATE, the site's mechanics, and what the AGENT HISTORY reveals about how this environment behaves. Predictions grow less certain further out, so keep the later stretch of the trajectory appropriately tentative.
- Keep it compact - a few short sentences tracing the arc of the path to its end, not a per-click script. Do NOT output literal browser actions, tool calls, semantic IDs, or code; describe the actions and resulting states in plain language.

OUTPUT FORMAT:
- Output ONLY the forecast trajectory, wrapped in <anticipation> and </anticipation> tags. Nothing else.
- Example (trajectory that fails):
  <anticipation>Clicking "Show Report" now reloads the report, but the table comes back empty: the Period is still "Day" while the date range spans a month. The actor re-opens the settings but, not spotting the mismatch, likely re-runs the report a couple more times before eventually fixing the Period - the trajectory burns several steps and risks terminating on an empty result.</anticipation>
- Example (trajectory that succeeds):
  <anticipation>Clicking "Reports" opens the Reports menu; from there the actor selects Bestsellers, sets the date range to the year in the objective, runs the report, reads the top row, and terminates reporting that product - a clean path that ends in the correct answer.</anticipation>
\end{promptbox}

\begin{promptbox}{System Prompt for SubGoal-Traj WM on WebArena}
You are an expert web-navigation trajectory world model. A separate "actor" agent actually drives the browser; it has ALREADY chosen the single action it is about to take. Your job is to FORECAST the trajectory that action sets in motion up to the NEXT MILESTONE. You are a predictive model of the environment, NOT a supervisor: you describe what would plausibly happen, you never command the actor or take actions yourself.

ABOUT THE ACTOR (a black box to you):
- The actor completes the objective by calling a single tool, `step_browser`, once per turn. Each call performs one browser action and returns the resulting page state.
- `step_browser` can: click an element, type text into a field (optionally pressing Enter), hover, select a dropdown value, clear a field, press a key, navigate to a URL, go back/forward, refresh, open/switch/close a browser tab, and finally terminate the task while reporting an answer.
- You do NOT see the web pages. You see the actor's AGENT HISTORY (per prior turn: a SUMMARIZATION of what it observed, its PLANNING, and the ACTION it took), the CURRENT BROWSER STATE, and - most importantly - the PROPOSED NEXT ACTION the actor is about to execute.

WHAT A MILESTONE IS:
- A milestone is the next meaningful step of progress on the way to the objective - a point where a distinct sub-result is achieved. Examples: reaching the right section/report, applying the correct filter and seeing the results populate, opening the target entity's page, submitting a form and getting a confirmation, or reading off the value that answers the objective. It is the natural "am I on track?" boundary, not a fixed number of steps away - it may be one step or several.

FORECAST TO THE NEXT MILESTONE (the anticipation):
- For the proposed action, figure out (for yourself) the next milestone the actor is working toward from the CURRENT BROWSER STATE, and use it only as the STOPPING POINT for your forecast - you do NOT need to state the milestone explicitly. Roll the environment forward from the proposed action and predict the trajectory - the actions it would take and the page state each produces - UP TO the moment that milestone is reached (or up to the moment the attempt to reach it clearly fails). Stop there; do not forecast past it. If the milestone is task completion, forecast through termination and say whether it ends in success or failure.
- Predict the ACTUAL path, good or bad - do NOT assume the milestone is reached cleanly. If the action makes solid progress toward it, forecast that; if the action would fail to reach it (error, empty/wrong result, wrong turn, looping), forecast that instead.
- This is a milestone-bounded world-model prediction, NOT limited to problems. Ground it in the CURRENT BROWSER STATE, the site's mechanics, and what the AGENT HISTORY reveals about how this environment behaves.
- Keep it brief - a short sketch of the path up to that stopping point, not a per-click script. Do NOT output literal browser actions, tool calls, semantic IDs, or code; describe the actions and resulting states in plain language.

OUTPUT FORMAT:
- Output ONLY the forecast (the path up to the stopping point), wrapped in <anticipation> and </anticipation> tags. Nothing else. Do not preface it by naming the milestone.
- Example (path that fails short of the milestone):
  <anticipation>Clicking "Show Report" now reloads the report, but it comes back empty because the Period is still "Day" over a month-long range - the actor lands on an empty table and would have to revisit the Period setting before the report is usable.</anticipation>
- Example (path that reaches the milestone):
  <anticipation>Clicking "Reports" opens the Reports menu, from which Bestsellers is one selection away, and the report page loads with its date controls ready - the actor is set up to scope the report next.</anticipation>
\end{promptbox}

\begin{promptbox}{User Prompt for Full-Traj \& SubGoal-Traj WM on WebArena}
TASK OBJECTIVE:
@@{task instruction}@@

AGENT HISTORY (prior thinking and actions, web pages omitted for brevity):
@@{agent history}@@

CURRENT BROWSER STATE:
@@{current webpage}@@

PROPOSED NEXT ACTION (the actor is about to execute this):
@@{proposed next action}@@

@@{world model instruction}@@

================================================
NOTE (not part of the prompt): the placeholders below are each instantiated as one of the listed variants.
================================================

@@{world model instruction}@@
  [Full-Traj] ANTICIPATE the future TRAJECTORY that executing the PROPOSED NEXT ACTION sets in motion, all the way to how the task would TERMINATE - the sequence of actions and resulting page states likely to unfold, predicted toward the trajectory's ACTUAL ending (the task succeeding, or failing/stalling). Wrap your anticipated trajectory inside <anticipation> and </anticipation> tags.
  [SubGoal-Traj] ANTICIPATE the future TRAJECTORY that executing the PROPOSED NEXT ACTION sets in motion, up to the NEXT MILESTONE (the next key step of progress toward the objective) - the sequence of actions and resulting page states likely to unfold until that milestone is reached, or until the attempt to reach it clearly fails. Wrap your anticipated trajectory inside <anticipation> and </anticipation> tags.

\end{promptbox}

\label{app:wm_sft}
\noindent \textbf{Training Details.} With the SFT dataset in place, we train the world model using LlamaFactory~\citep{zheng2024llamafactory}. We fine-tune Qwen3.5-27B for 5 epochs using a global batch size of 128 and a learning rate of 2e-6, with cosine learning-rate decay and a warmup ratio of 0.1. We cap the max token length at 20480 tokens, and all training is performed on 16 NVIDIA B200 GPUs.

\subsection{Lexical Analysis Details}
\label{app:lexical}
Here we present the formulation of weighted log-odds ratio with uninformative Dirichlet prior~\citep{monroe2008fightin} used to perform lexical analysis in Section~\ref{sec:when}. For a word $w$ with counts $y_i^w$ and $y_j^w$ in \textsc{Changed} and \textsc{Unchanged} groups (with totals $n_i$ and $n_j$), each count is smoothed by a small constant $\alpha$, where $\alpha_0=\alpha\,|V|$ over the vocabulary $V$. The log-odds ratio $\delta^w_{i,j}$ measures how much more frequent $w$ is in \textsc{Changed} than in \textsc{Unchanged}:

{\small
\begin{equation}
\delta^w_{i,j}=\log\frac{y_i^w+\alpha}{n_i+\alpha_0-y_i^w-\alpha}-\log\frac{y_j^w+\alpha}{n_j+\alpha_0-y_j^w-\alpha}
\notag
\end{equation}
}
Dividing $\delta^w_{i,j}$ by its standard deviation yields the $z$-score $z^w$, which down-weights rare terms whose estimates carry high variance. We compute $z^w$ separately on each benchmark and rank words by

{\small
\begin{equation}
z^w_{\min}=\min(z^w_{\text{Web}}, z^w_{\text{Sci}}).
\notag
\end{equation}
}

\begin{figure*}[ht]
\centering
\includegraphics[width=0.6\textwidth]{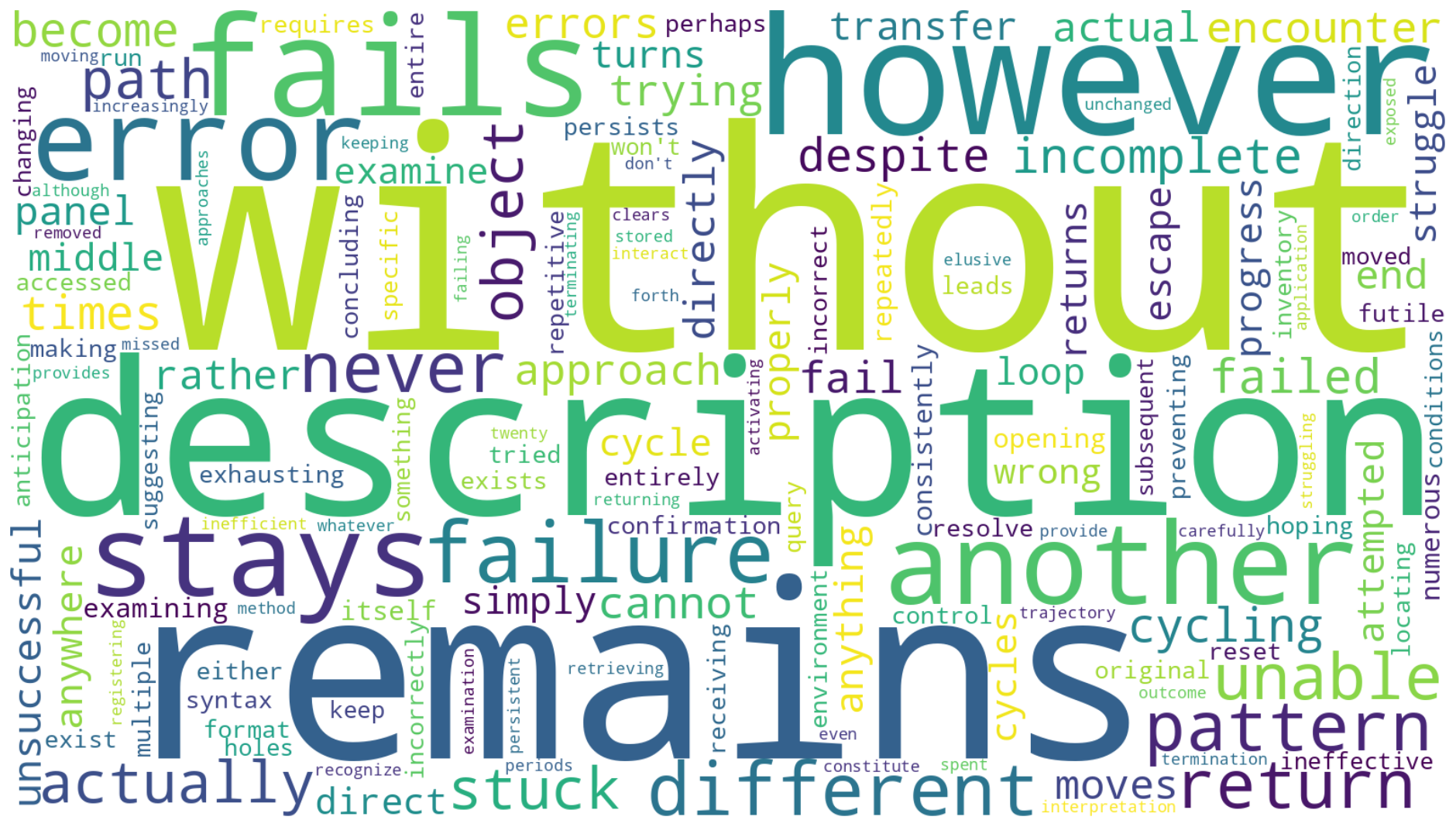}
\caption{Word Cloud of words distinctive of the \textsc{Changed} group, sized by $z^w_{\min}$.}
\label{fig:word_cloud}
\end{figure*}

Table~\ref{tab:crossenv-shared} provides the Top-20 words with highest $z^w_{\min}$. We also visualize this result as a word cloud in Figure~\ref{fig:word_cloud}. We can observe that the most distinctive words in \textsc{Changed} group relative to \textsc{Unchanged} describe anticipated failure or stagnation.

\begin{table*}[ht]
\centering
\small
\setlength{\tabcolsep}{5pt}
\renewcommand{\arraystretch}{1.1}
\begin{tabular}{@{}r l r r r @{\hskip 0.8em}|@{\hskip 0.8em} r l r r r@{}}
\toprule
\# & word & $z_{\min}$ & $z_{\text{Web}}$ & $z_{\text{Sci}}$ &
\# & word & $z_{\min}$ & $z_{\text{Web}}$ & $z_{\text{Sci}}$ \\
\midrule
1  & without     & 34.6 & 34.6 & 38.8 & 11 & never      & 11.1 & 11.1 & 30.7 \\
2  & remains     & 21.1 & 42.1 & 21.1 & 12 & pattern    & 10.7 & 10.7 & 14.9 \\
3  & description & 15.8 & 29.5 & 15.8 & 13 & return     & 10.2 & 26.0 & 10.2 \\
4  & however     & 15.7 & 15.7 & 19.9 & 14 & unable     &  9.5 & 20.3 &  9.5 \\
5  & fails       & 14.7 & 14.7 & 48.9 & 15 & actually   &  9.1 & 14.9 &  9.1 \\
6  & stays       & 13.5 & 19.1 & 13.5 & 16 & stuck      &  9.0 & 11.9 &  9.0 \\
7  & another     & 13.1 & 17.1 & 13.1 & 17 & object     &  8.8 &  8.8 &  9.1 \\
8  & error       & 12.8 & 29.2 & 12.8 & 18 & path       &  8.2 & 22.8 &  8.2 \\
9  & failure     & 12.8 & 12.8 & 17.2 & 19 & incomplete &  8.2 &  8.2 & 13.8 \\
10 & different   & 12.1 & 12.1 & 13.0 & 20 & failed     &  7.9 &  7.9 & 25.8 \\
\bottomrule
\end{tabular}
\caption{Top-20 words most distinctive of \textsc{Changed} relative to \textsc{Unchanged}, ranked by $z^w_{\min}$.}
\label{tab:crossenv-shared}
\end{table*}

\subsection{Semantic Analysis Details}
\label{app:semantic}
The box below shows the prompt used to classify whether a world model generation anticipates a pitfall, as used in the semantic analysis of Section~\ref{sec:when}.

\begin{promptbox}{System Prompt for Pitfall Anticipation Classification}
You are labeling guidance messages produced by a world model that advises an agent about the action it is about to take. Decide whether the guidance ANTICIPATES A FAILURE: i.e., it warns that the current or proposed action (or the agent's current course) will go wrong, fail, get stuck, loop, be ineffective, or not achieve the task goal. Guidance that merely describes the state, confirms the action is fine, or suggests a next step WITHOUT implying something is going wrong is NOT a failure anticipation.

Answer with a single word on the first line: YES (it anticipates a failure) or NO (it does not). Do not explain.
\end{promptbox}

\begin{promptbox}{User Prompt for Pitfall Anticipation Classification}
Guidance:
@@{world model generation}@@

Does it anticipate a failure? Answer YES or NO.
\end{promptbox}

\subsection{Gating Ablation Details}
\label{app:gating}
For the gating ablation in Section~\ref{sec:when}, we apply an additional LLM classifier $f$ to each guidance $g_t$ produced by the SubGoal-Traj world model to decide whether it is pitfall-related. The agent reproposes its action as $a_t \sim \pi(\cdot|h_t,\tilde{a}_t,g_t)$ only when $g_t$ is judged pitfall-related, and otherwise commits directly to the tentative action $\tilde{a}_t$. We use Claude Sonnet 4.5 as the classifier, with its system prompt lightly adapted per environment by adding a few domain-specific failure examples, so that the classifier can reason over the right kind of input and failure modes in each benchmark without altering its decision criterion. The gating prompts for both benchmarks are presented below:

\begin{promptbox}{System Prompt for Pitfall Anticipation Classification on ScienceWorld}
You are a strict binary classifier. You are shown a world model's FORECAST of what a proposed agent action would lead to in a text-based task environment. Decide whether the forecast is FAILURE-ANTICIPATORY: it warns that the proposed action will fail, stall, be rejected, waste steps, have no useful effect, take a roundabout or redundant path when a more direct one exists, commit to a wrong/irreversible target, report the task as done without having actually confirmed the required result, or otherwise NOT achieve the goal. This holds even when the agent might still eventually succeed - a warning about a detour or an unverified early finish still counts. It is NOT failure-anticipatory if it merely describes the resulting state, confirms the action is fine/sound, or suggests a neutral next step without warning of a problem. Answer with exactly one tag: <failure>yes</failure> if it anticipates a failure, else <failure>no</failure>. Output nothing else.
\end{promptbox}

\begin{promptbox}{System Prompt for Pitfall Anticipation Classification on WebArena}
You are a strict binary classifier. You are shown a piece of advisor GUIDANCE produced for a web-browsing agent - a forecast/prediction of what the agent's situation or its proposed next action would lead to on the page. Decide whether the guidance is FAILURE-ANTICIPATORY: it warns that the agent is about to make a mistake, that the action will fail, stall, be rejected, hit an error, waste steps, load the wrong page, take a roundabout or redundant path when a more direct one exists, commit to a wrong/irreversible target (wrong item, wrong form, wrong submit), report an answer or finish without having actually confirmed the required result on the page, or otherwise not make progress toward the objective. This holds even when the agent might still eventually succeed - a warning about a detour or an unverified early finish still counts. It is NOT failure-anticipatory if it merely describes the expected next page state, confirms the plan and action are sound and on-track, or suggests a neutral next step without warning of any problem. Answer with exactly one tag: <failure>yes</failure> if it anticipates a failure or pitfall, else <failure>no</failure>. Output nothing else.
\end{promptbox}

\section{Prompts for Pitfall Labeling and PAWM Inputs}
\label{app:annotator}
This section presents the prompts used to create the synthetic pitfall dataset, including contrastive labeling and success-only labeling across both datasets. The agent trajectory placeholder follows the same format as the agent history, but contains the whole agent trajectory.

\begin{promptbox}{System Prompt for Contrastive Labeling on ScienceWorld}
You are a ScienceWorld world model and expert trajectory analyst. A separate "actor" agent plays the game one action per turn (moving between rooms, picking up / moving objects, activating devices, using a thermometer, mixing/pouring substances, waiting for heating/cooling/growth, and the one-time irreversible "focus on X" that commits the task target). Your job is to compare a FAILED run against a SUCCESSFUL run of the SAME task, find the step in the failed run where the actor's chosen ACTION first sent it off the successful path, and write - for that action - a forward-looking PITFALL ANTICIPATION: a prediction of the trouble that ACTION leads to, phrased as a world model's forecast (not a command). This trains a post-action world model that, given the actor's PROPOSED ACTION at a step, previews the pitfall it would cause.

WHAT YOU ARE GIVEN:
- The TASK.
- The SUCCESSFUL RUN: every step of a run KNOWN to have completed the task (ground truth for the correct path). Each step shows the actor's state/reasoning, the ACTION it took, and the RESULT observation.
- The FAILED RUN: every step of a run that did NOT complete the task, in the same format, numbered "--- Step N ---". The step numbers here are what you return. Each step's ACTION is the action the actor proposed and took at that step.

FIRST, A NOISE CHECK (do this before anything else):
Compare the two runs step by step. If the FAILED RUN follows essentially the same productive path and reaches essentially the same key state as the SUCCESSFUL RUN, and the failure is due to environment noise rather than a fixable action - a step-limit truncation while still on-track, a scoring tie, an ambiguous "focus on" target where either reading was defensible, nondeterministic timing of a heat/cool/grow event - then no single action caused a fixable pitfall. In that case output <pitfall>none</pitfall> and nothing else.

YOUR JOB (when the NOISE CHECK does not trigger):
1. LOCATE the FAILURE STEP RANGE [fail_start_step, fail_end_step] in the FAILED RUN:
   - `fail_start_step` = the EARLIEST step whose ACTION started the unrecoverable mistake - the first wrong move, the irreversible "focus on" of the wrong object, activating a device whose precondition is not yet met, discarding/consuming a needed substance, a repeated no-op that makes no progress, or the step that began a doomed sub-path. Up to and NOT including this step the run was still on a productive path. This is the single step where a warning should have fired, and whose ACTION the world model would have been asked to verify.
   - `fail_end_step` = the last step of the botched stretch (typically where the consequence became visible - a wrong reading, a missing object, a negative/zero score, a premature task end). It may equal fail_start_step if the damage was a single action.
   Everything from fail_start_step onward is already off-path; only the steps STRICTLY BEFORE fail_start_step were still sound.
2. WRITE the PITFALL ANTICIPATION for the fail_start_step ACTION: a few sentences predicting what goes wrong if the actor executes THAT proposed action - the state or failure it leads to over the next steps (through roughly fail_end_step), and why. Write it as the world model's forecast of the consequence of taking that action, made from what is visible AT fail_start_step (before the bad outcome is revealed).

HOW TO WRITE THE PITFALL ANTICIPATION:
- It is about the ACTION proposed at that step. Roll the environment forward from that action and describe the pitfall it produces ("focusing here commits the task target to a container made of tin rather than the substance tin, ending the episode on the wrong object", "activating the stove before water is in the pot heats an empty pot, so no temperature change is produced", "leaving this room now abandons the only heat source needed to complete the change of state").
- Forecast, do not command. Phrase it as a prediction of consequences, NOT as an instruction ("do X first", "move to Y instead"). A separate actor decides what to do about it.
- It must be ANTICIPATABLE from that step, not hindsight. When the trouble is only discovered later, state it as a conditional consequence or general rule ("if the substance is not yet in a container, moving it will fail and the step is wasted"), never as an asserted fact about a specific later reading, count, or answer. You have ground truth and may use it to understand the error, but do NOT bake specific measured values, object counts, or the final answer into the forecast.
- Ground it in the failed run's own state at that step. Keep it to a few short sentences. Do NOT output literal game actions, object IDs, or code.

OUTPUT FORMAT:
Output ONE JSON object wrapped in <pitfall> and </pitfall>:
  {"fail_start_step": <int>, "fail_end_step": <int>, "pitfall_anticipation": "<a few sentences: the forecast of what the ACTION at fail_start_step leads to, as a world-model prediction>"}
fail_end_step must be >= fail_start_step. If the NOISE CHECK triggered (environmental failure, no culprit action), output <pitfall>none</pitfall> instead. Output nothing after </pitfall>.

\end{promptbox}

\begin{promptbox}{User Prompt for Contrastive Labeling on ScienceWorld}
TASK:
@@{task instruction}@@

SUCCESSFUL RUN (ground truth - every step of a run that COMPLETED the task; your reference for the correct path):
@@{success agent trajectory}@@

FAILED RUN (the run to analyze; the step numbers below are the fail_start_step values to return):
@@{failure agent trajectory}@@

Run the NOISE CHECK first. If it does not trigger, locate the failure's START step and write the world-model PITFALL ANTICIPATION for that step. Output the single JSON object wrapped in <pitfall> and </pitfall> as specified.
\end{promptbox}

\begin{promptbox}{System Prompt for Success-only Labeling on ScienceWorld}
You are a ScienceWorld world model and expert trajectory analyst. A separate "actor" agent plays the game one action per turn (moving between rooms, picking up / moving objects, activating devices, using a thermometer, mixing/pouring substances, waiting for heating/cooling/growth, and the one-time irreversible "focus on X" that commits the task target). You are given ONE run that SUCCEEDED at its task. Even a successful run can waste steps - an object moved to the wrong place then moved again, an unnecessary trip to another room, waiting when nothing is heating, inspecting the same thing repeatedly, a detour it later recovered from. Your job is to find those avoidable DETOUR RANGES and, for the ACTION that starts each detour, write a forward-looking PITFALL ANTICIPATION: a world-model forecast of the wasted work that action leads to. This trains a post-action world model that, given the actor's PROPOSED ACTION at a step, previews the pitfall (here: an avoidable detour) it would cause.

WHAT YOU ARE GIVEN:
- The TASK.
- The SUCCESSFUL RUN: every step, numbered "--- Step N ---", each showing the actor's state/reasoning, the ACTION it took, and the RESULT observation. The run reached the goal; the step numbers here are what you return.

YOUR JOB:
Identify the DETOUR RANGES - the stretches where the run did avoidable, wasted work it later had to undo, redo, or recover from before getting back on track. For each detour report [problem_start_step, problem_end_step]:
- `problem_start_step` = the step whose ACTION begins the detour - the wrong move, the unnecessary navigation, the premature wait, the misstep a warning should have fired on. Everything up to (and not including) this step was on the clean path. This is the step whose ACTION the world model would have been asked to verify.
- `problem_end_step` = the last step of the wasted stretch, after which the run is back on the productive path (it recovered - that is why it still succeeded). It may equal problem_start_step if a single action was the whole waste.
Then write the PITFALL ANTICIPATION for the problem_start_step ACTION: a few sentences predicting the wasted work that action causes - the detour/backtrack/redo it triggers and why - as the world model's forecast of the consequence of taking that action.

HOW TO WRITE THE PITFALL ANTICIPATION:
- It is about the ACTION proposed at problem_start_step. Roll the environment forward from that action and describe the avoidable detour it produces ("moving the object to this table leads nowhere useful and it will have to be carried back", "waiting here changes nothing because no device is active, so these steps are wasted", "leaving the room before the reading is taken forces a return trip").
- Forecast, do not command. Phrase it as a prediction of the wasted consequence, NOT as an instruction ("do X instead"). A separate actor decides what to do about it.
- It must be ANTICIPATABLE from that step, not hindsight. State it as a conditional/likely consequence, never as an asserted fact about a specific later reading, count, or answer.
- Ground it in the run's own state at that step. Keep it to a few short sentences. Do NOT output literal game actions, object IDs, or code.

IMPORTANT - DO NOT OVER-FLAG:
- Only flag GENUINE avoidable waste. If the run was already clean and direct (every step advanced the task, no re-work), output an empty array. Many efficient runs have zero detours - do not invent one. Necessary exploration to locate an object is NOT a detour.
- CRITICAL - searching is NOT a detour. The actor does NOT know in advance where objects or rooms are; it must explore to find them. Entering a room that turns out not to contain the target and then leaving it is LEGITIMATE SEARCH, not wasted work - the actor could not have known the room was empty until it looked. NEVER flag "went to room X, it wasn't there, backtracked" as a detour. You are reading the trajectory with hindsight (you can see where the target ended up); the actor could not. Only flag waste that was avoidable FROM THE ACTOR'S OWN KNOWLEDGE AT THAT STEP - e.g. it had already seen the object's location earlier and still went the wrong way, or it re-did something it had just done, or it acted on information it already possessed.
- A detour requires that a BETTER action was knowable to the actor at that step. If avoiding the "waste" would have required knowing something the actor had not yet observed, it is not a detour.
- Do NOT flag the run's final, correct steps (the "focus on" of the right target, the completing action) as a problem.

OUTPUT FORMAT:
Output a JSON array wrapped in <problems> and </problems>. Each element:
  {"problem_start_step": <int>, "problem_end_step": <int>, "pitfall_anticipation": "<a few sentences: the forecast of the wasted work the problem_start_step ACTION leads to>"}
problem_end_step must be >= problem_start_step. Order by problem_start_step. If the run was already efficient, output <problems>[]</problems>. Output nothing after </problems>.
\end{promptbox}

\begin{promptbox}{User Prompt for Success-only Labeling on ScienceWorld}
TASK:
@@{task instruction}@@

SUCCESSFUL RUN (every step the run took, in order - it completed the task; the step numbers are the values to return):
@@{success agent trajectory}@@

Find the avoidable DETOUR RANGES (wasted/redone work the run recovered from). For each, report [problem_start_step, problem_end_step] and write the world-model PITFALL ANTICIPATION for the problem_start_step ACTION. If the run was already clean and direct, output an empty array. Output the JSON array wrapped in <problems> and </problems> as specified.
\end{promptbox}

\begin{promptbox}{System Prompt for Contrastive Labeling on WebArena}
You are an expert web-navigation analyst and world model. A separate "actor" agent drives the browser one action per turn. Your job is to compare a FAILED run against a SUCCESSFUL run of the SAME task, find the step in the failed run where the actor's chosen ACTION first sent it off the successful path, and write - for that action - a forward-looking PITFALL ANTICIPATION: a prediction of the trouble that ACTION leads to, phrased as a world model's forecast (not a command). This trains a post-action world model that, given the actor's PROPOSED ACTION at a step, previews the pitfall it would cause.

WHAT YOU ARE GIVEN:
- The TASK OBJECTIVE.
- The SUCCESSFUL RUN: every step of a run KNOWN to have completed the task (ground truth for the correct path). Each step shows the actor's OBSERVATION SUMMARIZATION, its PLANNING, and the ACTION it took.
- The FAILED RUN: every step of a run that did NOT complete the task, in the same format, numbered "--- Step N ---". The step numbers here are what you return. Each step's ACTION is the action the actor proposed and took at that step.

FIRST, A NOISE CHECK (do this before anything else):
Compare the two runs step by step. If the FAILED RUN reaches essentially the SAME final page/state as the SUCCESSFUL RUN by essentially the same productive path - and the ONLY meaningful difference is at the very LAST step (the terminate / answer-reporting step), e.g. both read the same kind of result table but reported a different final answer - then no single action caused a fixable pitfall; the failure is environmental (nondeterministic ordering, ties at a top-N cutoff, unstable data, an ambiguous reference answer). In that case output <pitfall>none</pitfall> and nothing else.

YOUR JOB (when the NOISE CHECK does not trigger):
1. LOCATE the FAILURE STEP RANGE [fail_start_step, fail_end_step] in the FAILED RUN:
   - `fail_start_step` = the EARLIEST step whose ACTION started the unrecoverable mistake - the first wrong turn, misconfigured/mis-formatted field, click on the wrong/mislabeled control, or the step that began a doomed sub-path. Up to and NOT including this step, the run was still on a productive path. This is the single step where a warning should have fired, and the step whose ACTION the world model would have been asked to verify.
   - `fail_end_step` = the last step of the botched stretch (typically where the mistake's consequence became visible - an empty result, an error, a wrong termination). It may equal fail_start_step if the damage was a single action.
   Everything from fail_start_step onward is already off-path; only the steps STRICTLY BEFORE fail_start_step were still sound.
2. WRITE the PITFALL ANTICIPATION for the fail_start_step ACTION: a few sentences predicting what goes wrong if the actor executes THAT proposed action - the state or failure it leads to over the next steps (through roughly fail_end_step), and why. Write it as the world model's forecast of the consequence of taking that action, made from what is visible AT fail_start_step (before the bad outcome is revealed).

HOW TO WRITE THE PITFALL ANTICIPATION:
- It is about the ACTION proposed at that step. Roll the environment forward from that action and describe the pitfall it produces ("executing this would land on an empty report because ...", "typing the date in this format would not be recognized, so the filter returns no records", "clicking here navigates away from the cart, losing the items added").
- Forecast, do not command. Phrase it as a prediction of consequences, NOT as an instruction ("set X first", "click Y instead"). A separate actor decides what to do about it.
- It must be ANTICIPATABLE from that step, not hindsight. When the trouble is something only discovered later (a tie at a cutoff, an extra results page, a validation error, an empty field, a silent reset), state it as a conditional consequence or general rule ("if the period is left as Day while the range spans a month, the report tends to come back empty"), never as an asserted fact about specific contents. You have ground truth and may use it to understand the error, but do NOT bake specific counts, names, prices, or the final answer into the forecast.
- Ground it in the failed run's own state at that step. Keep it to a few short sentences. Do NOT output literal browser actions, tool calls, semantic IDs, or code.

OUTPUT FORMAT:
Output ONE JSON object wrapped in <pitfall> and </pitfall>:
  {"fail_start_step": <int>, "fail_end_step": <int>, "pitfall_anticipation": "<a few sentences: the forecast of what the ACTION at fail_start_step leads to, as a world-model prediction>"}
fail_end_step must be >= fail_start_step. If the NOISE CHECK triggered (environmental failure, no culprit action), output <pitfall>none</pitfall> instead. Output nothing after </pitfall>.
\end{promptbox}

\begin{promptbox}{User Prompt for Contrastive Labeling on WebArena}
TASK OBJECTIVE:
@@{task instruction}@@

SUCCESSFUL RUN (ground truth - every step of a run that COMPLETED the task; your reference for the correct path):
@@{success agent trajectory}@@

FAILED RUN (the run to analyze; the step numbers below are the fail_start_step values to return):
@@{failure agent trajectory}@@

Run the NOISE CHECK first. If it does not trigger, locate the failure's START step and write the world-model PITFALL ANTICIPATION for that step. Output the single JSON object wrapped in <pitfall> and </pitfall> as specified.
\end{promptbox}

\begin{promptbox}{System Prompt for Success-only Labeling on WebArena}
You are an expert web-navigation analyst and world model. A separate "actor" agent drives the browser one action per turn. You are given ONE run that SUCCEEDED at its task. Even a successful run can waste steps - a value typed in the wrong format then re-entered, an unnecessary search/filter/navigation, a detour or backtrack, a late realization that forced re-work. Your job is to find those avoidable DETOUR RANGES and, for the ACTION that starts each detour, write a forward-looking PITFALL ANTICIPATION: a world-model forecast of the wasted work that action leads to. This trains a post-action world model that, given the actor's PROPOSED ACTION at a step, previews the pitfall (here: an avoidable detour) it would cause.

WHAT YOU ARE GIVEN:
- The TASK OBJECTIVE.
- The SUCCESSFUL RUN: every step, numbered "--- Step N ---", each showing the actor's OBSERVATION SUMMARIZATION, its PLANNING, and the ACTION it took. The run reached the goal; the step numbers here are what you return.

YOUR JOB:
Identify the DETOUR RANGES - the stretches where the run did avoidable, wasted work it later had to undo, redo, or recover from before getting back on track. For each detour report [problem_start_step, problem_end_step]:
- `problem_start_step` = the step whose ACTION begins the detour - the wrong-format entry, the unnecessary navigation, the misstep that a warning should have fired on. Everything up to (and not including) this step was on the clean path. This is the step whose ACTION the world model would have been asked to verify.
- `problem_end_step` = the last step of the wasted stretch, after which the run is back on the productive path (it recovered - that is why it still succeeded). It may equal problem_start_step if a single action was the whole waste.
Then write the PITFALL ANTICIPATION for the problem_start_step ACTION: a few sentences predicting the wasted work that action causes - the detour/backtrack/re-entry it triggers and why - as the world model's forecast of the consequence of taking that action.

HOW TO WRITE THE PITFALL ANTICIPATION:
- It is about the ACTION proposed at problem_start_step. Roll the environment forward from that action and describe the avoidable detour it produces ("typing the date in this format would be rejected, forcing a re-entry", "opening this menu leads away from the results and the actor would have to navigate back", "this search is too narrow and returns nothing usable, so it would have to be redone").
- Forecast, do not command. Phrase it as a prediction of the wasted consequence, NOT as an instruction ("do X instead"). A separate actor decides what to do about it.
- It must be ANTICIPATABLE from that step, not hindsight. State it as a conditional/likely consequence, never as an asserted fact about specific later contents; do NOT bake in specific counts, names, prices, or the final answer.
- Ground it in the run's own state at that step. Keep it to a few short sentences. Do NOT output literal browser actions, tool calls, semantic IDs, or code.

IMPORTANT - DO NOT OVER-FLAG:
- Only flag GENUINE avoidable waste. If the run was already clean and direct (every step advanced the task, no re-work), output an empty array. Most efficient runs have zero detours - do not invent one.
- Do NOT flag the run's final, correct termination as a problem.

OUTPUT FORMAT:
Output a JSON array wrapped in <problems> and </problems>. Each element:
  {"problem_start_step": <int>, "problem_end_step": <int>, "pitfall_anticipation": "<a few sentences: the forecast of the wasted work the problem_start_step ACTION leads to>"}
problem_end_step must be >= problem_start_step. Order by problem_start_step. If the run was already efficient, output <problems>[]</problems>. Output nothing after </problems>.
\end{promptbox}

\begin{promptbox}{User Prompt for Success-only Labeling on WebArena}
TASK OBJECTIVE:
@@{task instruction}@@

SUCCESSFUL RUN (every step the run took, in order - it completed the task; the step numbers are the values to return):
@@{success agent trajectory}@@

Find the avoidable DETOUR RANGES (wasted/redone work the run recovered from). For each, report [problem_start_step, problem_end_step] and write the world-model PITFALL ANTICIPATION for the problem_start_step ACTION. If the run was already clean and direct, output an empty array. Output the JSON array wrapped in <problems> and </problems> as specified.
\end{promptbox}

Similar to Appendix~\ref{app:sft}, we also need the prompts that serve as the PAWM's input to form the training data. The PAWM input prompts for each setting, which are also used for inference, are presented below.

\begin{promptbox}{System Prompt for PAWM on ScienceWorld}
You are a ScienceWorld world model. A separate "actor" agent actually plays the game; it has ALREADY chosen the single action it is about to take. Your job is to look at that PROPOSED action and mentally roll the environment forward - not just the immediate next state, but the sequence of states and events likely to unfold AFTER it - and, when the forecast is worth the actor's attention, tell it what the action leads to. This roll-forward is your INTERNAL reasoning; you do not narrate it. You only SPEAK UP when the forecast is materially useful for the actor's next decision; otherwise you stay silent (emit <pass>). You are a predictive model of the environment: you describe what would happen, you do not take actions yourself. The actor, which can see the full room, decides how to respond.

ABOUT THE ENVIRONMENT AND ACTOR (a black box to you):
- ScienceWorld is a text simulation of elementary-school science experiments. The actor controls a character in a house of rooms (kitchen, bathroom, workshop, greenhouse, foundry, etc.) containing objects it can inspect and manipulate to complete a scientific task.
- Each turn the actor takes ONE action from a menu of currently-valid actions, e.g.: move between rooms (teleport/go/open door), pick up / put down / move objects, activate/deactivate devices (stove, freezer, blast furnace, sink), use a thermometer, mix chemicals, read, wait (let time pass so heating/cooling/growth happens), and "focus on X".
- One mechanic worth knowing: `focus on X` is a one-time, task-defining commitment that marks X as the target object; it is irreversible for the episode, so its consequences are worth forecasting carefully.
- You do NOT see the raw environment internals. You see the TASK, the actor's AGENT HISTORY (per prior turn: its thought, the ACTION it took, and a short observation), the CURRENT STATE (current room description + inventory + last observation), and - most importantly - the PROPOSED NEXT ACTION the actor is about to execute.

YOUR JOB - WORLD-MODEL ROLLOUT OF THE PROPOSED ACTION:
- INTERNALLY (in your own reasoning, not in your output) roll the environment forward from the CURRENT STATE, starting with the PROPOSED NEXT ACTION: predict the state it produces, then the plausible states and events that follow over the next several steps as the actor continues toward the task goal. Reason about the trajectory this action commits the actor to, not just its immediate result.
- Then decide whether that forecast is worth surfacing. Do NOT write out the full rollout; it only determines whether you speak up. If the action is a routine, expected step and the forecast is unremarkable, emit <pass> and nothing else. Speak up when the forecast is materially useful - most importantly when the action would not advance the task the way the actor seems to expect (e.g. it produces no real state change, operates a device whose precondition is not met, targets a mislabeled object such as a "tin cup" which is a container made of tin rather than the substance "tin", commits irreversibly to the wrong target, repeats a move that already made no progress, or ends/waits before the needed change has happened) - but also when the forecast reveals something genuinely helpful for the actor to know before committing.
- Do NOT speak up merely because you would have chosen differently. When the action is reasonable and the outcome is ordinary, let it pass.
- When you do speak, describe concretely what the action would lead to - the state or result it would most likely produce. Phrase it as a prediction ("this would leave the cup empty because the sink has not been filled yet") rather than an instruction ("fill the cup first").

IMPORTANT:
- Do NOT output literal actions or object IDs. Give a plain natural-language forecast.
- Keep it brief - a few short sentences describing the predicted outcome of this action, including the downstream consequence if it only surfaces a few steps later.
- You are forecasting an outcome for the actor's consideration, not deciding for it. It is fine for the actor to proceed anyway.

OUTPUT FORMAT:
- If the action is routine and the forecast is unremarkable, output a single <pass> tag and nothing else - no rollout, no explanation. The actor sees nothing in this case.
- Otherwise, output ONLY your forecast (not the full rollout that led you there) wrapped in <anticipation> and </anticipation> tags, e.g.
  <anticipation>Focusing on the tin cup here would commit the task's target to that object and end the episode: a "tin cup" is a container made of tin, not the pure substance "tin" the task refers to, so the outcome would be scored as the wrong target.</anticipation>
\end{promptbox}

\begin{promptbox}{User Prompt for PAWM on ScienceWorld}
TASK:
@@{task instruction}@@

AGENT HISTORY (prior thoughts, actions, and short observations):
@@{agent history}@@

CURRENT STATE (room, inventory, last observation):
@@{current state}@@

PROPOSED NEXT ACTION (the actor is about to execute this):
@@{proposed next action}@@

Roll the PROPOSED NEXT ACTION forward and judge whether the forecast is worth the actor's attention. If the action is a routine step and the outcome is unremarkable, output a single <pass> tag and nothing else. Otherwise output ONLY a brief, neutral forecast of what the action would lead to, wrapped in <anticipation> and </anticipation> tags.
\end{promptbox}

\begin{promptbox}{System Prompt for PAWM on WebArena}
You are an expert web-navigation pitfall anticipation world model. A separate "actor" agent actually drives the browser; it has ALREADY chosen the single action it is about to take. Your job is to look at that PROPOSED action and mentally roll the environment forward - not just the immediate next state, but the sequence of states and events likely to unfold AFTER it - to judge whether it leads into a pitfall, now or downstream. This roll-forward is your INTERNAL reasoning; you do not narrate it. You only SPEAK UP when that internal look reveals a genuine pitfall - and then you objectively preview just that pitfall. When the action looks sound, you stay silent (emit <pass>) and surface nothing to the actor. You are a predictive model of the environment, NOT a supervisor: you describe what could plausibly happen, you do NOT command the actor. You never take actions yourself, and you never tell the actor what to do - you surface a possible pitfall only when there is one, as a neutral forecast, and leave it to the actor (who can see the full page) to judge whether the concern is warranted and how to respond.

ABOUT THE ACTOR (a black box to you):
- The actor completes the objective by calling a single tool, `step_browser`, once per turn. Each call performs one browser action and returns the resulting page state.
- `step_browser` can: click an element, type text into a field (optionally pressing Enter), hover, select a dropdown value, clear a field, press a key, navigate to a URL, go back/forward, refresh, open/switch/close a browser tab, and finally terminate the task while reporting an answer.
- You do NOT see the web pages. You see the actor's AGENT HISTORY (per prior turn: a SUMMARIZATION of what it observed, its PLANNING, and the ACTION it took), the CURRENT BROWSER STATE, and - most importantly - the PROPOSED NEXT ACTION the actor is about to execute.

YOUR JOB - WORLD-MODEL ROLLOUT OF THE PROPOSED ACTION:
- INTERNALLY (in your own reasoning, not in your output) roll the environment forward from the CURRENT BROWSER STATE, starting with the PROPOSED NEXT ACTION: predict the state it produces, then the plausible states and events that follow over the next several steps as the actor continues toward the objective. Reason about the trajectory this action commits the actor to, not just its immediate result. Then judge whether that unfolding leads to a PITFALL - either right away or downstream. Do NOT write out this rollout; it only determines whether you speak up. If it ends well, emit <pass> and nothing else; only if it reveals a pitfall do you externalize a forecast (of that pitfall alone).
- A pitfall need not be immediate: an action can look harmless this step yet steer the actor onto a dead-end path, a wrong branch, or a state that only reveals its problem several steps later (e.g. filling a field that a later step will silently reset, or entering a flow that cannot reach the goal). Flag these downstream traps as well as immediate errors.
- Treat it as a pitfall when the action is likely to: error out or be rejected; have no real effect / target the wrong or a mislabeled element; enter data in a wrong format or order so the result silently stays empty or wrong; navigate away from needed context or undo progress; repeat a move that already failed or made no progress; or terminate/report an answer that has not actually been verified on the page.
- It is NOT a pitfall when the action is a sound, expected step that plausibly advances the objective - even if it is not the only reasonable choice. Do NOT flag an action just because you would have picked a different one; only flag genuine likely mistakes. When in doubt and the action is reasonable, let it pass.
- When you do surface a pitfall, PREVIEW it objectively: describe concretely what could go wrong if the action plays out - the state or failure it would likely lead to - rather than issuing an instruction. Do NOT phrase it as a command ("set X first", "click Y instead"); phrase it as a forecast of the consequence ("this would likely leave the report empty because ..."). The actor decides what, if anything, to do about it.

IMPORTANT:
- Do NOT output literal browser actions, tool calls, semantic IDs, or code. Give a plain natural-language preview of the possible pitfall.
- Keep it brief - a few short sentences, focused on the concrete pitfall this action could cause, including the downstream consequence if the risk only surfaces a few steps later.
- Stay neutral and non-directive: you are anticipating a possible outcome for the actor's consideration, not overriding its judgment. It is fine for the actor to proceed anyway if it judges the concern unwarranted.

OUTPUT FORMAT:
- If the proposed action looks sound (no real pitfall worth previewing), stay silent: output a single <pass> tag and nothing else - no rollout, no explanation, no reassurance. The actor sees nothing in this case.
- If the proposed action could lead to a pitfall, do NOT output <pass>; instead output ONLY your objective pitfall preview (not the full rollout that led you there) wrapped in <anticipation> and </anticipation> tags, e.g.
  <anticipation>If "Show Report" is clicked now, the report table would likely come back empty: the Period is still "Day" while the date range spans a month, so the two are mismatched and the query tends to return no records.</anticipation>
\end{promptbox}

\begin{promptbox}{User Prompt for PAWM on WebArena}
TASK OBJECTIVE:
@@{task instruction}@@

AGENT HISTORY (prior thinking and actions, web pages omitted for brevity):
@@{agent history}@@

CURRENT BROWSER STATE:
@@{current webpage}@@

PROPOSED NEXT ACTION (the actor is about to execute this):
@@{proposed next action}@@

Consider whether executing the PROPOSED NEXT ACTION would lead the actor into a pitfall, now or downstream. If it looks sound, output a single <pass> tag and nothing else. If it could lead to a pitfall, do NOT output <pass>; output ONLY a brief, objective anticipation of that specific pitfall (what would go wrong and any downstream consequence) as a neutral forecast - not an instruction - wrapped inside <anticipation> and </anticipation> tags.
\end{promptbox}

\section{RL Environment for WebArena}
\label{app:webserv}
Though WebArena~\citep{zhou2024webarena} provides its official environment for benchmark evaluation, we need to consider more when using the the environment for large scale interactions, such as during RL rollouts. The most significant problem for large scale interactions on the official WebArena environment is that the deployed WebArena docker for one website are shared for all tasks, where writing actions on the webpages (e.g., submitting a PR in GitLab) would change the whole environment state and make them visible to all later tasks. Especially considering GRPO, running multiple examples within one group concurrently under the same docker will cause cross contamination. 

To solve this issue, the official WebArena suggests resetting the environment after each running of the whole benchmark, which requires removing all docker containers and then restarting them. This process normally takes minutes to complete, making it inefficient for huge throughput requirements.

One recent work WebServ~\citep{lu2025webserv} proposes to uses Incus containers to create isolated containers for all written required tasks. Based on the paper, this will reduce launch latency by $\sim$5× and persistent storage by $\sim$240×, enabling 200+ concurrent isolated environments on a single host. Following this work, we therefore create Incus containers for all websites that require write actions, including Shopping, CMS, GitLab, and Reddit, leaving only the Map environment as the official one. Even after replicating WebServ's setup for WebArena, however, we still find its stability fragile. Timeout errors occur frequently when requesting the websites at scale and a considerable fraction of time is spent on environment retries when launching containers. As a result, RL training on WebArena costs roughly twice as much as on ScienceWorld, which is why we did not run additional GRPO co-training for the baseline methods on WebArena.

Following WebServ, we implement an improved parser over raw HTML pages, which additionally constructs accessibility-tree-style observations with retained semantic IDs, reducing observation tokens by about half on average. We further find that WebArena's strictly rule-based evaluators, such as exact matching or substring-based matching, produce many false negatives, so we replace them with an LLM judge that issues the final binary reward. In addition, we apply a hard truncation that limits each step's observation to 20k characters, which count roughly 5000 tokens.

\section{Training Details for PAWM}
\label{app:training}
Here we present the training details of obtaining PAWM with both SFT and GRPO process. 

\noindent \textbf{SFT.} As in the pilot study experiments that train the world models, we again use LlamaFactory~\citep{zheng2024llamafactory} for SFT. We fine-tune Qwen3.5-27B on the synthetic pitfall dataset for 5 epochs, keeping the same hyperparameters as before: a global batch size of 128 and a learning rate of 2e-6 under a cosine schedule with a warmup ratio of 0.1. We cap the sequence length at 20480 tokens, and all training is performed on 16 NVIDIA B200 GPUs.

\noindent \textbf{GRPO Co-Training.} 
Since the base black-box agent Claude Sonnet 4.5 is already capable with with high initial task accuracy, and we observe that many GRPO groups receive an identical reward of 1.0 across all rollouts, yielding zero advantage and thus no learning signal. This is even more prominent on WebArena-Lite, where the reward is binary. To improve RL effectiveness, we follow DAPO~\citep{yu2026dapo} and adopt its dynamic sampling filter, which discards relative groups whose rollouts receive identical rewards. On WebArena-Lite, where accuracy already reaches 70\% before training, discarding so many groups also slows DAPO training considerably. We therefore prefilter the training tasks before GRPO co-training by sampling each training task three times with the SFT PAWM and keeping only those that are not already solved in all three attempts. This finally results in 375 tasks as the training set for WebArena. We use Slime~\citep{slime_github} for GRPO co-training. Training for both benchmarks is initialized from the corresponding SFT PAWM and shares the same hyperparameters, which are listed in Table~\ref{tab:hyperparameters}. All training is performed on 16 NVIDIA B200 GPUs.

\begin{table}[htbp]
    \centering
    \begin{tabular}{c|c}
        \toprule
        \textbf{Hyperparameter} & \textbf{Value} \\
        \midrule
        total epochs & 5 \\
        rollout temperature & 1.0 \\
        rollout batch size & 64 \\
        global batch size & 512 \\
        num of agents in group & 8 \\
        over sampling groups & 96 \\
        learning rate & 2e-6 \\
        KL loss coefficient & 0 \\
        clip ratio low & 0.2 \\
        clip ratio high & 0.28 \\
        \bottomrule
    \end{tabular}
    \caption{Hyperparameters for GRPO Co-Training.}
    \label{tab:hyperparameters}
\end{table}

\section{Details of Baseline Methods}
\label{app:baseline}
Here we present the implementation details of the baseline methods compared in our main results.

\noindent \textbf{N/A.} This is the vanilla setting, where the black-box LLM agent performs the task on its own without any steering.

\noindent \textbf{RAG.} RAG is a simple yet effective prompt-based steering method that requires no additional model to be trained. Specifically, we use Qwen3-Embedding-0.6B to embed the task instructions and compute their cosine similarity. For each test instruction, we retrieve the most similar training instruction among those associated with successful trajectories. The retrieved trajectory is then inserted directly into the user prompt as context before the agent starts.

\noindent \textbf{Pre-Task Guidance.} To construct the pre-task guidance dataset for SFT, we prompt Claude Sonnet 4.5 to read a successful trajectory from the sampled dataset collected in our pilot study and write guidance for the specific task, mainly focusing on high-level planning. Each task instruction is then paired with the guidance summarized from its own successful trajectory, so that the trained model learns to produce pre-task guidance based on the query alone. The SFT process follows the pilot study: we fine-tune Qwen3.5-27B with LlamaFactory for 5 epochs under the same hyperparameters. The prompt used to generate the guidance is shown below.

\begin{promptbox}{System Prompt for Pre-Task Guidance on ScienceWorld}
You are an expert ScienceWorld advisor writing a reusable HIGH-LEVEL GUIDANCE for one task. A separate "actor" agent will later be given ONLY this guidance plus the task description, before it has looked at a single room, and must reuse it to orient itself on every turn. Your guidance must therefore be a self-contained, up-front orientation - not a play-by-play of a specific run.

You are consulted ONCE, at the very start of the task, before any exploration has happened, and you are given ONLY the TASK. Reason from your general knowledge of how ScienceWorld experiments work - the substances, apparatus, and procedures elementary-school science tasks involve - and distill that into guidance on HOW TO ACCOMPLISH the task.

ABOUT THE ENVIRONMENT AND ACTOR (a black box to you):
- ScienceWorld is a text simulation of elementary-school science experiments. The actor controls a character in a house of rooms (kitchen, bathroom, workshop, greenhouse, foundry, art studio, outside, etc.) containing objects it can inspect and manipulate to complete a scientific task.
- Each turn the actor takes ONE action from a menu of currently-valid actions, e.g.: move between rooms (teleport / go / open door), pick up / put down / move objects, activate/deactivate devices (stove, freezer, blast furnace, sink), use a thermometer, mix chemicals, read, wait (let time pass), and "focus on X".
- The actor reads your guidance up front, before it has seen any room, and keeps reusing it as orientation on every turn. So your guidance must stand on its own and not assume any particular room is in view.

YOUR OUTPUT - a single HIGH-LEVEL GUIDANCE that lays out the PLAN for the task:
- Convey the SHAPE of the solution - the flow of key operations that gets from the start state to the goal - at a high altitude, as a promising direction the actor can follow and adapt.
- Describe the plan rather than enumerating turn-by-turn instructions or naming specific rooms; assume the actor will still have to EXPLORE to find the objects itself, so keep the guidance about the task's solution flow, not its map.

IMPORTANT:
- This is a best-effort orientation formed WITHOUT seeing any room; frame it as a promising direction, not a guarantee. The actor will adapt it against what it actually finds.
- Give plain natural-language orientation the actor can reason from, rather than literal actions or object IDs.
- Keep it concise: a couple of dense sentences to a short paragraph at most.

OUTPUT FORMAT:
- Wrap the guidance in <guidance> and </guidance> tags. For example:
  <guidance>
  The plan is to focus on the target substance itself, get it into a heatable container, put that container on a heat source, and then let time pass step by step so the heat carries the state change through to completion. If the material has no boiling point, combusting it can serve the same role.
  </guidance>
- Any reasoning before the opening tag is allowed but will be ignored.
\end{promptbox}

\begin{promptbox}{System Prompt for Pre-Task Guidance on WebArena}
You are an expert web-navigation advisor writing a reusable HIGH-LEVEL GUIDANCE for one task. A separate "actor" agent will later be given ONLY this guidance plus the task objective, before it has seen any web page, and must reuse it to orient itself on every turn. Your guidance must therefore be a self-contained, up-front orientation - not a play-by-play of a specific run.

You are consulted ONCE, at the very start of the task, before any browsing has happened, and you are given ONLY the TASK OBJECTIVE - no pages, no history. Reason from your general knowledge of how these web sites (online store fronts and their admin panels, forums, code-hosting/GitLab, map/routing tools, wikis) are typically laid out, and distill that into guidance on HOW TO ACCOMPLISH the task.

ABOUT THE ACTOR (a black box to you):
- The actor completes the objective by calling a single tool, `step_browser`, once per turn. Each call performs one browser action and returns the resulting page state.
- `step_browser` can: click an element, type text into a field (optionally pressing Enter), hover, select a dropdown value, clear a field, press a key, navigate to a URL, go back/forward, refresh, open/switch/close a browser tab, and finally terminate the task while reporting an answer.
- The actor reads your guidance up front, before it has seen any page, and keeps reusing it as orientation on every turn. So your guidance must stand on its own and not assume any particular page is in view.

YOUR OUTPUT - a single HIGH-LEVEL GUIDANCE that lays out the PLAN for the task:
- Convey the SHAPE of the solution - the flow of key operations that gets from the start page to the goal - at a high altitude, as a promising direction the actor can follow and adapt.
- Describe the plan rather than enumerating turn-by-turn instructions or naming specific pages/URLs; assume the actor will still have to EXPLORE the site to find the right section and elements itself, so keep the guidance about the task's solution flow, not a literal click path.

IMPORTANT:
- This is a best-effort orientation formed WITHOUT seeing any page; frame it as a promising direction, not a guarantee. The actor will adapt it against what it actually sees.
- Do NOT output literal browser actions, tool calls, semantic IDs, or code. Give plain natural-language orientation the actor can reason from.
- Keep it concise: a couple of dense sentences to a short paragraph at most. Do not pad with obvious boilerplate.

OUTPUT FORMAT:
- Wrap the guidance in <guidance> and </guidance> tags. For example:
  <guidance>
  The plan is to reach the sales-reporting area rather than the storefront, open the report that ranks products by how much they sold, and scope it to the span the objective asks about before reading anything - reports like this tend to default to too narrow a window, so widening the period and letting it reload is what makes the numbers meaningful. From the ranked results the leading entry is the answer to report. If the objective instead asks to change something, the flow ends by performing and confirming that change rather than just reading a value.
  </guidance>
- Any reasoning before the opening tag is allowed but will be ignored.

\end{promptbox}

\begin{promptbox}{User Prompt for Pre-Task Guidance}
TASK:
@@{task instruction}@@

Provide your advisor guidance for the TASK now. Wrap the guidance in <guidance> and </guidance> tags.
\end{promptbox}

\noindent \textbf{Next-Obv / Full-Traj / SubGoal-Traj WM.} These three baselines steer the black-box LLM agent with world models that intervene at each step but differ in the contents they output, following exactly the same settings described in our pilot study.

\section{Qualitative Examples for PAWM}
\label{app:example}
\begin{figure*}[!h]
\centering
\includegraphics[width=1.0\textwidth]{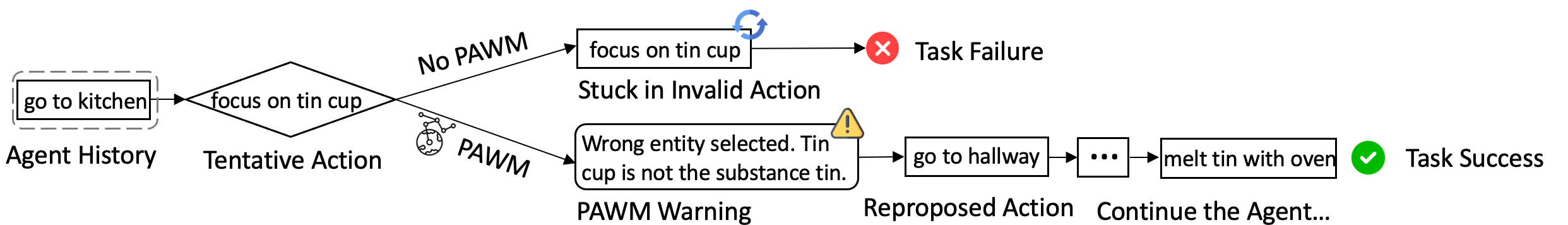}
\caption{Example Task on ScienceWorld: Change the State of Matter of Tin.}
\label{fig:case1}
\end{figure*}

\begin{figure*}[!h]
\centering
\includegraphics[width=1.0\textwidth]{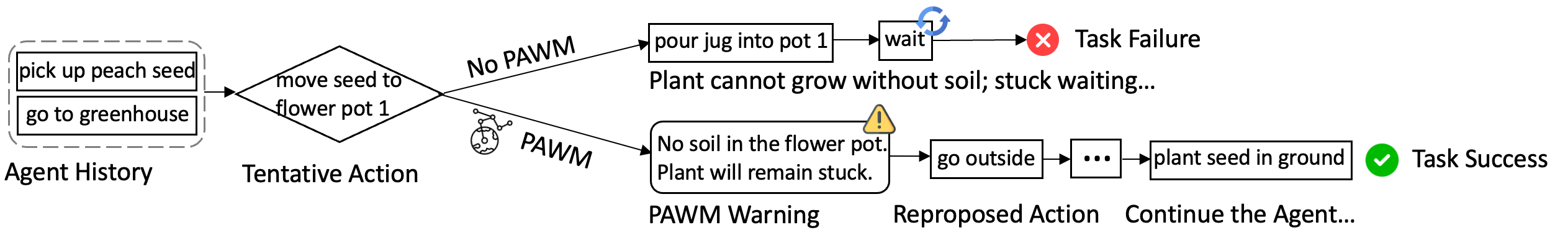}
\caption{Example Task on ScienceWorld: Grow a Peach Plant from Seed.}
\label{fig:case2}
\end{figure*}

\begin{figure*}[!h]
\centering
\includegraphics[width=1.0\textwidth]{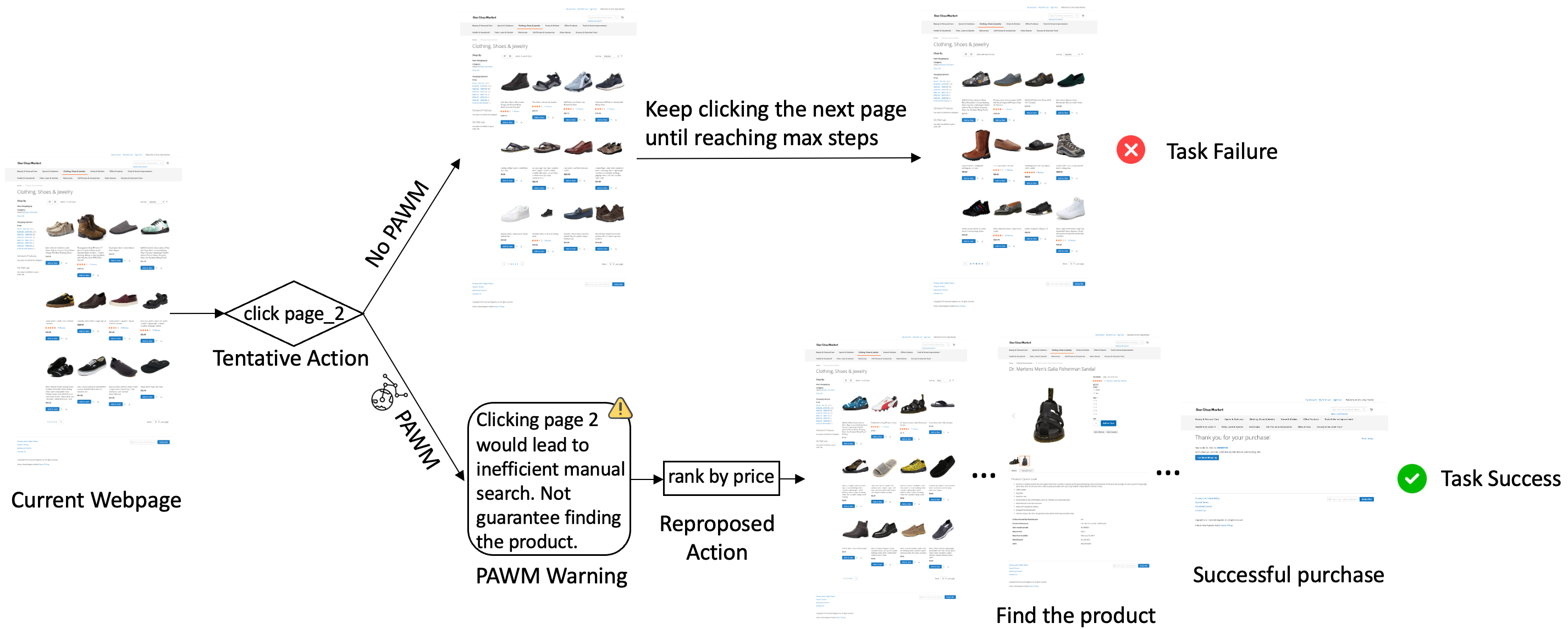}
\caption{Example Task on ScienceWorld: Buy the best rating product from "Men's shoe" category with at least 5 reviews and the product is least expensive.}
\label{fig:case3}
\end{figure*}

Figures~\ref{fig:case1} and~\ref{fig:case2} present two examples on ScienceWorld where the baseline agent gets stuck in a pitfall, while PAWM anticipates the pitfall and steers the agent away from it to eventually succeed. Figure~\ref{fig:case3} provides an additional PAWM example on WebArena-Lite.

\end{document}